\documentclass[10pt,journal,compsoc]{IEEEtran}
\ifCLASSOPTIONcompsoc
  \usepackage[nocompress]{cite}
\else
  \usepackage{cite}
\fi

\ifCLASSINFOpdf
\else
\fi

\usepackage{amsmath,amsfonts}
\usepackage{algorithmic}
\usepackage{algorithm}
\usepackage{array}
\usepackage[caption=false,font=normalsize,labelfont=sf,textfont=sf]{subfig}
\usepackage{textcomp}
\usepackage{stfloats}
\usepackage{url}
\usepackage{verbatim}
\usepackage{graphicx}
\usepackage{cite}

\usepackage[capitalize]{cleveref}
\usepackage{multirow}
\usepackage[table,xcdraw,dvipsnames]{xcolor}
\usepackage{threeparttable, tablefootnote}
\usepackage{arydshln}

\begin{document}

\title{MS3D++: Ensemble of Experts for Multi-Source Unsupervised Domain Adaptation in 3D Object Detection}

\author{Darren Tsai, Julie Stephany Berrio, Mao Shan, Eduardo Nebot and Stewart Worrall%
\thanks{This work was supported by Transport for NSW, University of Sydney (Australian Centre for Robotics), iMOVE Australia, and ARC LIEF grant LE200100049 Whopping Volta GPU Cluster - Transforming Artificial Intelligence Research. Additionally, this research was supported partially by the Australian Government through the ARC DP funding scheme (project DP220102019) (Corresponding author: Darren Tsai.)}
\thanks{The authors are with the Australian Centre For Robotics (ACFR) at the University of Sydney (NSW, Australia). E-mails: {\small{\{d.tsai, j.berrio, m.shan, e.nebot, s.worrall}\}@acfr.usyd.edu.au}}%
}

\IEEEtitleabstractindextext{%
\begin{abstract}
Deploying 3D detectors in unfamiliar domains has been demonstrated to result in a significant 70-90\% drop in detection rate due to variations in lidar, geography, or weather from their training dataset. This domain gap leads to missing detections for densely observed objects, misaligned confidence scores, and increased high-confidence false positives, rendering the detector highly unreliable. To address this, we introduce MS3D++, a self-training framework for multi-source unsupervised domain adaptation in 3D object detection. MS3D++ generates high-quality pseudo-labels, allowing 3D detectors to achieve high performance on a range of lidar types, regardless of their density. Our approach effectively fuses predictions of an ensemble of multi-frame pre-trained detectors from different source domains to improve domain generalization. We subsequently refine predictions temporally to ensure temporal consistency in box localization and object classification. Furthermore, we present an in-depth study into the performance and idiosyncrasies of various 3D detector components in a cross-domain context, providing valuable insights for improved cross-domain detector ensembling. Experimental results on Waymo, nuScenes and Lyft demonstrate that detectors trained with MS3D++ pseudo-labels achieve state-of-the-art performance, comparable to training with human-annotated labels in Bird's Eye View (BEV) evaluation for both low and high density lidar. Code is available at \url{https://github.com/darrenjkt/MS3D}.
\end{abstract}

}

\maketitle

\IEEEdisplaynontitleabstractindextext

%
\IEEEpeerreviewmaketitle

\IEEEraisesectionheading{\section{Introduction}\label{sec:introduction}}
\IEEEPARstart{I}{n} recent years, 3D object detection has made remarkable progress, benefiting from advancements in detection architectures and the availability of large-scale autonomous vehicle datasets \cite{sun2020waymo,geiger2012kitti,caesar2020nuscenes,woven2019lyft,alibeigi2023zenseact}. However, detectors often encounter significant challenges when deployed in an unfamiliar domain with drastic differences from the training data due to domain shift. This shift is caused by variations in lidar types, weather conditions, and geographical locations and has shown to lead to a drastic drop of up to 70-90\% in detection performance. For example, deploying a pre-trained detector to upgraded lidar sensors may suffer from domain shift, necessitating the collection and annotation of massive datasets specific to each new lidar. Additionally, in real-world scenarios, mass-produced robots and vehicles typically use lidars with lower point density compared to large-scale public datasets commonly used for 3D detector training. Models deployed on such platforms may also suffer from poor performance due to the domain gap between training and testing domains. In both cases, collecting and manually annotating data for re-training detectors can be costly and impractical, highlighting the need to leverage existing labeled data for detector adaptation. We refer to this task as Unsupervised Domain Adaptation (UDA).

\begin{figure}[t]
  \centering
  \includegraphics[width=0.99\linewidth]{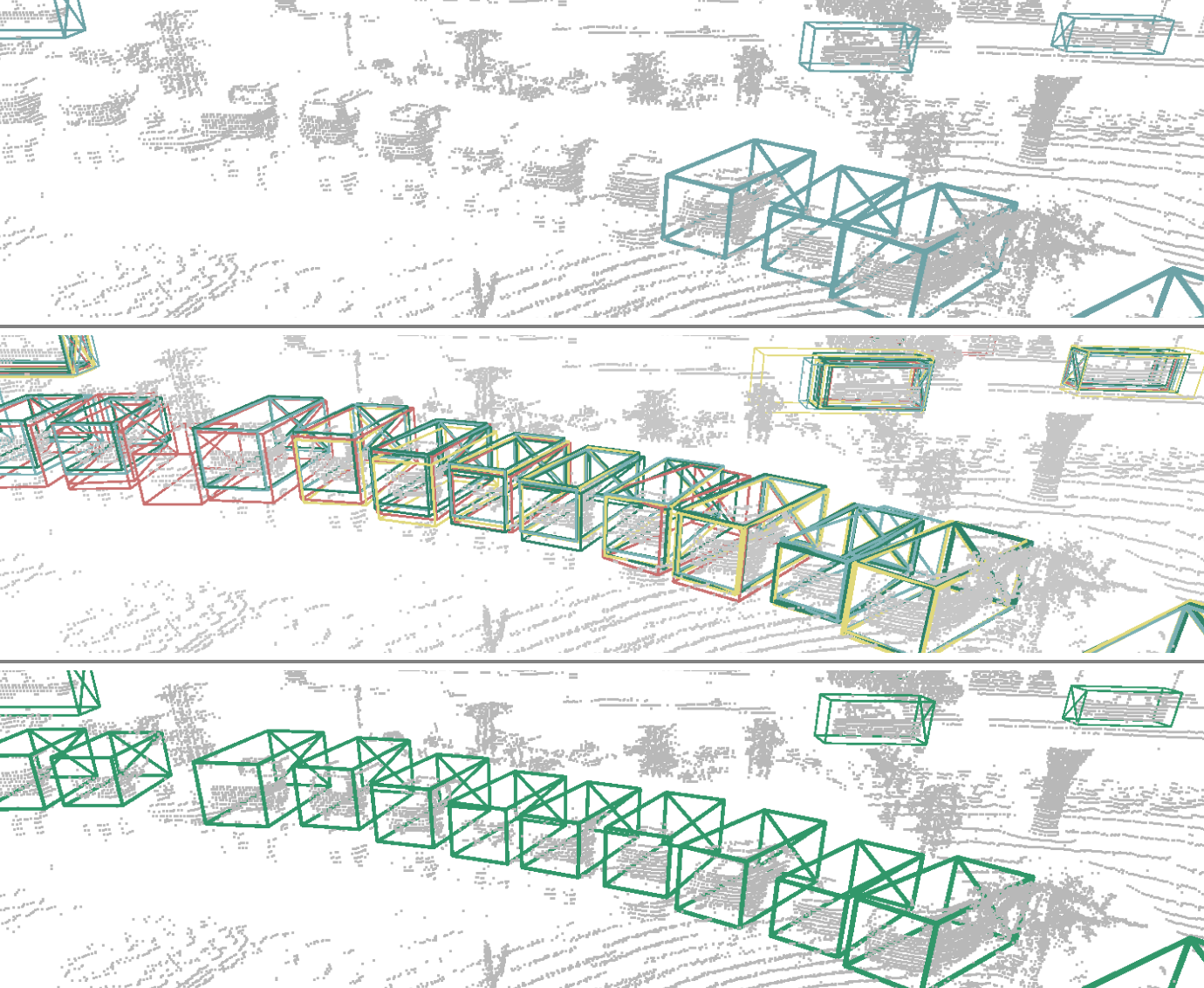}
  \caption{Domain generalization for unseen target domains can be improved by ensembling multiple pre-trained detectors from various source domains. Figure shows detector predictions on a point cloud from the Waymo dataset \cite{sun2020waymo} in rainy weather. Top: PV-RCNN++ \cite{shi2023pv++} trained on Lyft\cite{woven2019lyft}, Middle: Ensemble of PV-RCNN++ and VoxelRCNN\cite{deng2020voxelrcnn} trained on both Lyft and nuScenes\cite{caesar2020nuscenes}, Bottom: Fused detections with our Kernel-density estimation Box Fusion (KBF).}
  \label{fig:ensemble}
  \vspace{-4mm}
\end{figure}

\begin{figure*}
  \centering
  \includegraphics[width=0.99\linewidth]{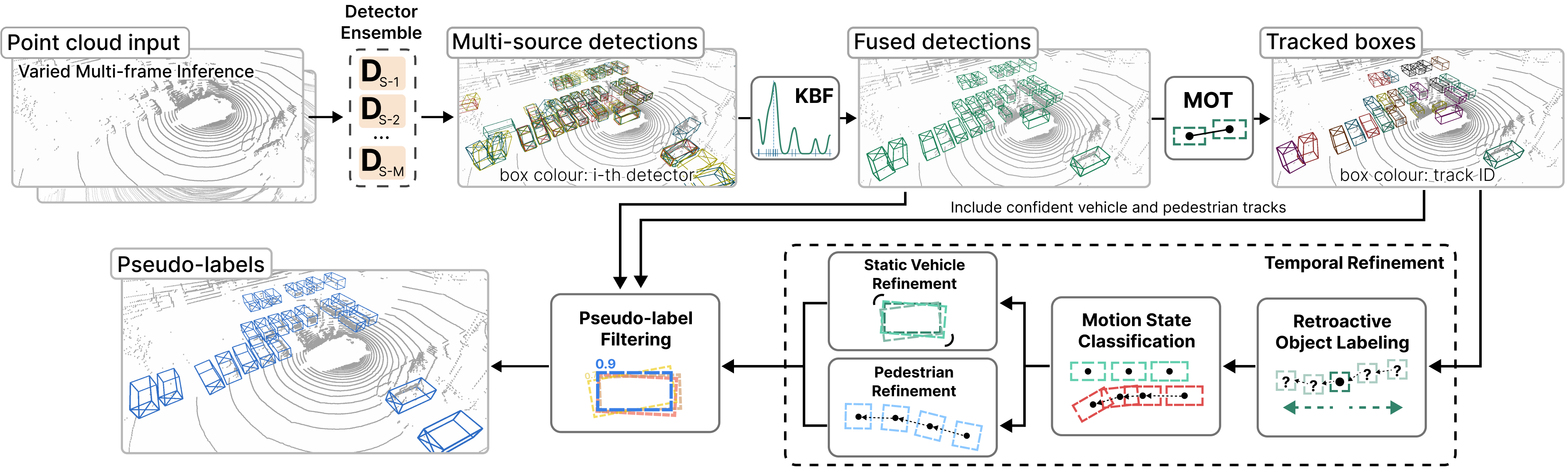}
  \caption{Our \textbf{multi-source self-training framework, MS3D++}. Given a set of $M$ pre-trained, multi-frame 3D detectors from multiple source domains $\textbf{D}_{\text{S,m}}$ where $\text{m}=1,2,...,\text{M}$, we generate predictions for a varying number of accumulated point cloud frames which are fused with KBF and tracked. Our temporal refinement uses object characteristics to ascertain class labels and refine bounding box localization. The set of pseudo-labels is iteratively improved with each update, utilizing the re-trained detectors from the previous round in its ensemble.}
  \label{fig:ms3d++_framework}
  \vspace{-3mm}
\end{figure*}

Unsupervised domain adaptation is a challenging problem, as it requires addressing domain shift without access to labeled data in the target domain. Self-training has been proven to be an effective solution for domain adaptation \cite{yang2021st3d,yihan2021learning,tsai2023ms3d,saltori2020sf,luo2021mlcnet,peng2023cl3d}. By generating pseudo-labels for the target domain, we can effectively bridge the domain gap and adapt detectors to new environments. However, existing self-training approaches typically focus on adapting a single detector from one source to a new target domain \cite{yang2021st3d,luo2021mlcnet,you2022exploiting}. With this approach, the source dataset and detector choice can significantly impact the domain adaptation performance. For example, adapting a Waymo-trained \cite{sun2020waymo} detector to KITTI \cite{geiger2012kitti} achieves a far higher performance than adapting a nuScenes-trained \cite{caesar2020nuscenes} detector to KITTI \cite{hu2023density, yang2021st3d} due to Waymo and KITTI both having 64-beam lidars. In practice, with the ongoing development of novel lidar types, evaluating and identifying the optimal source dataset or detector for a particular scan pattern may be challenging due to the lack of labeled data on target domains. To address this, we propose MS3D++, a multi-source self-training framework illustrated in \cref{fig:ms3d++_framework}. Through multi-source detector ensembling and temporal refinement, we demonstrate that 3D detectors trained with MS3D++ achieve state-of-the-art performance on both low and high density lidars, even comparable to the BEV detection of detectors trained with human-annotated labels.

MS3D++ is built upon three fundamental observations. Firstly, we recognize that different detector architectures trained on different point cloud datasets possess distinct expertise  \cite{tsai2023ms3d}. For instance, a Waymo-trained detector with smaller voxel sizes may be better suited for a newly released high density lidar compared to a nuScenes-trained detector. Using an ensemble of experts allows us to obtain a more robust set of initial pseudo-labels on an unfamiliar target domain, as illustrated in \cref{fig:ensemble}. Secondly, many works have demonstrated the benefits of short-sequence point cloud accumulation when trained and tested on the same lidar setup \cite{caesar2020nuscenes, yang20213dman}. Our study reveals that multi-frame pre-trained detectors can also bring significant improvements when tested in a cross-domain setting. This allows us to achieve increased detection of sparse and far-range objects, substantially boosting the performance of our domain adaptation baseline. Lastly, to address the issue of unreliable detector confidence scores in a cross-domain context, we use temporal information which has been proven effective in 3D box refinement \cite{qi2021offboard,ma2023detzero,fan2023oncedetected} for a supervised setting. Leveraging temporal information allows us to accurately distinguish true positive class predictions and accurately refine box localization.

In real-world scenarios, Multi-Source Domain Adaptation (MSDA) is a highly feasible approach for adaptation to a new target domain. The availability of multiple large-scale datasets and detector architectures effectively forms a vast database of detector and source domain pairings that can be used for domain adaptation. MS3D++ demonstrates that ensembling detectors from various source domains with different architectures can lead to improved domain generalization for a given target domain. The versatility of MS3D++ lies in its capacity to tailor the 3D detector ensemble to include detectors with different architectures, source domains, hyperparameters, or class-specific detectors to a specific target domain making it an adaptable solution for a wide range of lidars and scenes. Moreover, MS3D++ presents a simple approach to domain adaptation by using the generated set of labels as a replacement for human-annotated labels in a supervised setting. This enables easy integration with any 3D detector architecture and data augmentation technique, making MS3D++ highly practical for adapting existing and future 3D detectors while preserving their inference runtime. This sets MS3D++ apart from other UDA works that require detector architecture modifications \cite{zhang2021srdan}, are limited to detectors with Bird's Eye View (BEV) feature extraction \cite{yihan2021learning,zhang2023uni3d,hu2023density}, or involve additional inference steps \cite{tsai2022see, tsai2023viewer}.

In summary, this paper presents MS3D++, and significantly expands upon our previous work, MS3D \cite{tsai2023ms3d} by (1) providing a quantitative analysis of 3D detector components' cross-domain performance across diverse datasets, which leads to improved ensembling performance, (2) extending MS3D to devise a robust pseudo-labeling strategy for pedestrians, (3) introducing Varied Multi-Frame Inference to effectively use multi-frame pre-trained detectors for both vehicles and pedestrians in a cross-domain setting, (4) introducing a multi-stage self-training framework to increase pseudo-label recall while preserving precision, and (5) surpassing the performance of our previous work, therefore establishing a new state-of-the-art for 3D detectors on unseen target domains without requiring any manually labeled data. 

The rest of this paper is organized as follows: In \cref{sec:related_work}, we give an overview of relevant research. \cref{sec:quantifying_cross_domain_detection} presents a quantitative analysis of the cross-domain performance of 3D detectors, offering key insights into differences across datasets that contribute to the domain gap. Moving on to \cref{sec:ms3d++}, we introduce MS3D++, informed by observations drawn from our cross-domain analysis with extensive evaluation in \cref{sec:experiments}. We release our code, pseudo-labels and reproducible models to facilitate future research in this field.

\section{Related Work}
\label{sec:related_work}
\subsection{3D Object Detection}
3D object detection aims to localize and classify objects in a point cloud. In \cref{sec:quantifying_cross_domain_detection}, we conduct experiments to study how different components of 3D detectors perform in cross-domain settings. In this section, we therefore aim to provide a preliminary overview of existing point cloud representations and detection heads used in 3D detector architectures and offer insights that are relevant to our cross-domain study. 

\subsubsection{Point cloud representations} Due to the sparse and irregular nature of point clouds, various approaches aim to efficiently and accurately encode spatial information from raw point clouds. The representation of point clouds can be largely categorized into three main categories: voxel-based, point-based and hybrid representations. Voxel-based\cite{yan2018second,deng2020voxelrcnn,zhou2018voxelnet,chen2023voxelnext} partition the point cloud into voxels that can be processed by mature Convolutional Neural Network (CNN) architectures and robust 2D detection heads. This representation offers regularity and efficient feature encoding of multi-scale features due to its strong memory locality, which also facilitates fast neighbour querying \cite{deng2020voxelrcnn, liu2019pointcnn}. However,  high resolution (small voxel sizes) is required to preserve location information which is computationally expensive for long range detection. At lower resolutions, multiple points may be assigned to the same voxel, resulting in a loss of structural details. 

Point-based representations \cite{zhang2022iassd,yang20203dssd,shi2019pointrcnn} on the other hand, often process raw points with PointNet \cite{qi2017pointnet, qi2017pointnet++}, using symmetric functions for point feature extraction to tackle the unordered nature of point cloud data. These methods use sampled points as key points and aggregate local regions (i.e., grouping) around them to encode precise position information. However, constructing local region sets by finding neighboring points often involves costly nearest neighbor searches due to irregular point distribution in 3D space. 

Hybrid representations\cite{shi2020pv, yang2019std, shi2020parta2, noh2021hvpr,chen2017mv3d,guan2022m3detr} explore combining multiple representations of the point cloud. For example, MV3D \cite{chen2017mv3d} and MVF \cite{zhou2020mvf} proposes to combine BEV and range view, whilst others explore fusing points, voxels and BEV with a transformer \cite{guan2022m3detr}. A prominent state-of-the-art approach, PV-RCNN \cite{shi2020pv}, along with its successor\cite{shi2023pv++}, focus on combining the strengths of both point and voxel-based methods to achieve efficient multi-scale feature encoding while preserving precise location information. This approach alleviates the grouping inefficiency of the point-based methods by sampling fewer keypoints, and aggregating neighbouring voxel features instead of points. 

\subsubsection{Detection head} The detection head is responsible for processing feature maps into bounding box and object class predictions. These can be classified into two categories: anchor-based and anchor-free. Anchor-based detection heads such as the Region Proposal Network (RPN) \cite{yan2018second, he2017maskrcnn} place predefined anchors at uniform intervals on the feature map. The model is thereafter trained to regress an offset from these anchor box priors to the final bounding box. Anchor sizes are often selected based on the mean sizes of objects in a training dataset. The benefit of anchor-based methods are that axis-aligned boxes can be a very strong prior for regressing vehicles due to the nature of driving scenes where vehicles are often axis-aligned. The disadvantage however, is that objects come in various sizes and orientations, making it challenging to robustly regress accurate object size and orientation \cite{yin2021centerpoint}. Anchor-free \cite{yin2021centerpoint,shi2019pointrcnn, hu2022afdetv2, zhang2022iassd} approaches focus on predicting foreground points \cite{shi2019pointrcnn, zhang2022iassd} or centroids \cite{yin2021centerpoint} of objects for regression of the bounding boxes. This addresses the bias of anchor-based methods towards axis-aligned predictions, as well as removes the burden of choosing IOU thresholds for the anchor target assignment for new classes or datasets.

\subsection{Unsupervised Domain Adaptation}
Unsupervised Domain Adaptation (UDA) focuses on adapting a model trained on a labeled dataset (source domain), to a new, unlabeled dataset (target domain). UDA works can be categorised into domain-invariant representation\cite{yi2021complete, tsai2022see, tsai2023viewer}, adversarial methods\cite{qin2019generatively, saito2019strong}, self-training \cite{yang2021st3d,yihan2021learning,tsai2023ms3d,saltori2020sf,luo2021mlcnet,peng2023cl3d}, amongst others. Within the UDA task, there are a few distinct settings. We refer to cross-domain adaptation as a general category of works that focus on adapting detectors from one source domain to an unfamiliar target domain \cite{tsai2023ms3d,yang2021st3d}. This is also referred to as single-source to single-target UDA. In this setting, ST3D \cite{yang2021st3d,yang2022st3d++} proposed a self-training framework with a thresholding method that selects high-confidence detections as pseudo-labels, however this can lead to the discarding of correct labels with low-confidence. Some works address specific components of the domain shift, such as SRDAN \cite{zhang2021srdan} which focuses on time-of-day, weather and synthetic-to-real domain adaptation through learning a domain-invariant representation of object geometry. \cite{wang2020train} highlighted that the main domain gap issue for adaptation of US-based datasets \cite{sun2020waymo,caesar2020nuscenes} to German-based datasets \cite{geiger2012kitti} is in the sizes of the cars. 

Multi-target UDA, or more generally, domain generalization, aims to train solely on the source data and annotations, to learn a representation that enables more robust performance on a wide range of domains. SEE \cite{tsai2022see, tsai2023viewer} is a multi-target adaptation work that proposes to transform point clouds into a unified scan pattern representation that is generalizable across multiple types of lidars. Source-free UDA is another category of works \cite{tsai2023ms3d,saltori2020sf,ahmed2021msda} where only the pre-trained detectors and unlabeled target data are available. In a real-world setting, source data may not be available for a variety of reasons, such as privacy and security. It may also not be practical to transfer or store large-scale datasets on different platforms for adaptation \cite{ahmed2021msda}. SF-UDA \cite{saltori2020sf} is a source-free self-training approach that uses a tracker to re-score pseudo-labels. 

Multi-source domain adaptation (MSDA) approaches use a bag of source domains, where each domain is correlated to the target domain by different amounts. This task involves not only robustly fusing the knowledge of multiple models but simultaneously preventing the chance of negative transfer \cite{ahmed2021msda}. In UDA for 3D detection, there are only two works that explore MSDA. Uni3D \cite{zhang2023uni3d} proposes to train a detector on multiple source datasets with statistics-level alignment in the feature space to learn data-agnostic features and prevent negative transfer. Our prior work, MS3D \cite{tsai2023ms3d} is a source-free, MSDA approach that ensembles multiple pre-trained detectors, and temporal refinement for pseudo-labeling within a self-training framework. MS3D focuses on the real-world scenario where pre-trained detectors are widely available for use, and outperforms all existing approaches, achieving state-of-the-art performance for 3D detectors on new target domains. 

\begin{figure*}
  \centering
  \includegraphics[width=0.99\linewidth]{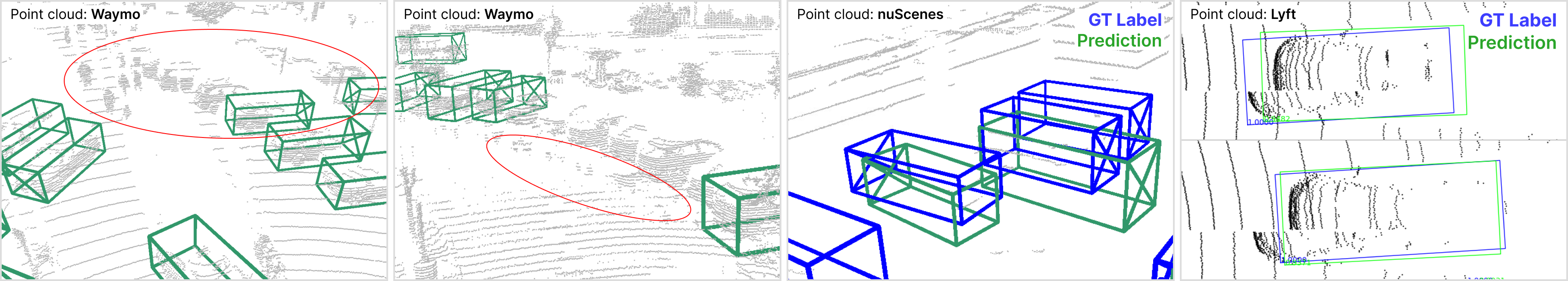}
  \caption{3D Detectors tested on cross-domain data encounter various issues. From left to right: (1) Scan pattern discrepancy causes densely observed vehicles and pedestrians to be missed, (2) Weather artefacts such as reflections on the ground due to rain, appear different across various scan patterns, causing missed detections or false positives, (3) Sparse lidar is challenging for box height estimation, (4) Poor (or lack of) motion compensation for the ego-vehicle causes artefacts, inaccurate labels, and therefore, prediction ``errors". In the image, the upper predicted box may not meet the IoU=0.7 threshold for a true positive detection. Predictions for image (1-4) were made with PV-RCNN++ (centerhead) and trained on nuScenes (1), Lyft (2) and Waymo (3-4). We elaborate on these in \cref{sec:study_cross_domain_results}.} 
  \label{fig:dataset_domain_gap}
  \vspace{-3mm}
\end{figure*}

\section{Quantifying Cross-Domain 3D object detection}
\label{sec:quantifying_cross_domain_detection}
Cross-domain 3D object detection is the task of applying a 3D detector, trained on a labeled source dataset, to a separate target dataset where factors such as sensor setup and geography may vary significantly. This domain shift between training and testing domains, often leads to low-quality predictions and poses challenges, shown in \cref{fig:dataset_domain_gap}. 

Many existing works propose methods to address cross-domain adaptation, however the study of cross-domain performance without adaptation (aka. ``Source-Only" or ``Direct Transfer") of 3D detectors is often overlooked. We believe that a comprehensive understanding of the baseline cross-domain performance and domain gap idiosyncrasies will enhance the community's ability to obtain higher domain adaptation performance, and overall, better generalization to any given target domain. For example, most UDA approaches adapt a 32-beam (nuScenes) to a 64-beam lidar with a detector trained on 1-frame of nuScenes as the pre-trained model \cite{yang2021st3d,hu2023density,chen2023revisiting}. However, we demonstrate in \cref{sec:train_with_accum} that simply pre-training a nuScenes detector with 10 accumulated frames increases the detection on Waymo point clouds by $+21 \text{AP}_{3D}$. We show in \cref{tab:multiframe_comparison_with_st3d} that a multi-frame pre-trained Waymo detector can even outperform a prominent self-training UDA work, ST3D \cite{yang2021st3d,yang2022st3d++}, when testing on nuScenes. Overall, we demonstrate that multi-frame pre-training presents a stronger baseline performance on various target domains without requiring complicated source-target alignment approaches. 

In this section, we seek to analyse and quantify the performance of various detector components in the context of a cross-domain setting to study the idiosyncrasies of 3D detectors. Furthermore, we use these insights to better incorporate pre-trained 3D detectors in MS3D++.

\setlength{\tabcolsep}{0.2em} 
\begin{table}[t]
\caption{Overview of datasets.}
\label{tab:datasets}
\centering
\begin{tabular}{ccccccc} 
\hline
Dataset  & Lidar Beams     & Size     & Rain & Night & Country  \\ 
\hline
nuScenes & 1$\times$32            & 34,149         & Yes  & Yes    & USA, Singapore \\
Waymo    & 1$\times$64 $+$ 4$\times$200    &  198,068     & Yes  & Yes    &  USA \\
Lyft     & 1$\times$40 or 64 $+$ 2$\times$40 & 22,680       & No   & No    & USA   \\
\hline
\end{tabular}
\vspace{-2mm}
\end{table}

\subsection{Experiment Setup}
\label{sec:quantifying_cross_domain_exp_setup}

\subsubsection{3D Detector Selection} To investigate the performance of various point cloud representations in the cross-domain setting, we select five detectors that are representative of the various point cloud representations and detection heads. For point cloud representation we select VoxelRCNN\cite{deng2020voxelrcnn} (voxel-based), IA-SSD \cite{zhang2022iassd} (point-based) and PV-RCNN++\cite{shi2023pv++} (hybrid). For detection heads, we select Anchorhead \cite{yan2018second} (aka. RPN, anchor-based) and Centerhead \cite{yin2021centerpoint} (anchor-free) as our representative methods. Except for IA-SSD, we select two-stage 3D detectors, PV-RCNN and VoxelRCNN, due to their superior performance over single-stage detectors. For PV-RCNN and VoxelRCNN, we conduct experiments with both anchor and centerhead. In summary, we conduct this study on five detectors: (1) PV-RCNN with Anchorhead (PV-A), (2) PV-RCNN with Centerhead (PV-C), (3) VoxelRCNN with Anchorhead (VX-A), (4) VoxelRCNN with Centerhead (VX-C), and (5) IA-SSD. Following existing cross-domain adaptation works \cite{yang2021st3d,tsai2023ms3d}, each voxel-based detector for every dataset is trained with a voxel size of [0.1,0.1,0.15]m, detection range of $[-75,75]$m for x,y and $[-2,4]$m for z-axis, and we shift the lidar frame to the ground plane. We train detectors on point clouds $\{P_i \in \mathbf{R}^{n_i \times C}\}$, $i=1,2,...,N$ with the $i$-th frame point cloud $P_i$ consisting of $n_i$ points with $C$ channels for each point, for $N$ total frames. For single frame 3D detection, each point in $P_i$ has channels $C=3$ consisting of the $XYZ$ point coordinates. For multi-frame 3D detection, each point has $C=4$ channels corresponding to $XYZT$ where $T$ is the timestamp which we elaborate on in \cref{sec:train_with_accum}. We follow default settings from \cite{openpcdet2020} for other hyperparameters.

\subsubsection{Dataset Selection and Evaluation} We focus on three popular datasets for this study: (1) Waymo Open Dataset \cite{sun2020waymo} with a 64-beam lidar, (2) nuScenes \cite{caesar2020nuscenes} with a 32-beam lidar, and (3) Lyft \cite{woven2019lyft} with a 40-beam and 64-beam lidar. These are large-scale datasets, summarised in \cref{tab:datasets}, with over 20,000 frames of labeled data and containing variations in scan pattern, geography, and weather. We note that the frame rate of each of these datasets are different: Waymo is given at 10Hz, nuScenes has 2hz key frames with 20Hz sweeps, and Lyft is at 5Hz. As a result, the historical accumulation of 10 frames in nuScenes (0.45s) is not an equivalent time period to 10 frames in Waymo (0.9s). We report the 3D Average Precision ($\text{AP}_{\text{3D}}$) for IoU at 0.7 using Waymo's range evaluation of L2/AP for each table on ``Vehicle" and ``Pedestrian" classes. ``Vehicle" evaluation in Lyft and nuScenes includes the car, truck and bus subcategories. 

\begin{table}[]
\centering
\setlength{\tabcolsep}{0.3em} 
\caption{Multi-frame vs single-frame detectors on cross-domain data. We show $\text{AP}_\text{3D}$ gain ($\text{AP}_{\text{multi}}-\text{AP}_{\text{single}}$) on official validation datasets.}
\label{tab:analysis_multiframe}
\begin{tabular}{l|c|c|cc|cc}
\hline
                                    &                                   &                                   & \multicolumn{2}{c|}{$\text{AP}_\text{3D}$ Gain}                  & \multicolumn{2}{c}{$\text{AP}_\text{3D}$ Gain (\%)}                                             \\ \cline{4-7} 
\multirow{-2}{*}{Detector} & \multirow{-2}{*}{Source} & \multirow{-2}{*}{Target} & Veh.                & Ped.                & Veh.                          & Ped.                             \\ \hline
                                    &                                   & nuScenes                          & {\color[HTML]{036400} +3.1}  & {\color[HTML]{036400} +2.8}  & {\color[HTML]{036400} 20.6\%}             & {\color[HTML]{036400} 57.1\%}             \\
                                    & \multirow{-2}{*}{Waymo}           & Lyft                              & {\color[HTML]{036400} +0.6}  & {\color[HTML]{036400} +2.9}  & {\color[HTML]{036400} 1.3\%}              & {\color[HTML]{036400} 8.6\%}              \\ \cline{2-7} 
                                    &                                   & Lyft                              & {\color[HTML]{036400} +12.8} & {\color[HTML]{036400} +8.2}  & {\color[HTML]{036400} 58.4\%}             & {\color[HTML]{036400} 92.2\%}             \\
                                    & \multirow{-2}{*}{nuScenes}        & Waymo                             & {\color[HTML]{036400} +21.0} & {\color[HTML]{036400} +3.3}  & {\color[HTML]{036400} 141.9\%}            & {\color[HTML]{036400} 154.9\%}            \\ \cline{2-7} 
                                    &                                   & nuScenes                          & {\color[HTML]{036400} +1.3}  & {\color[HTML]{036400} +2.8}  & {\color[HTML]{036400} 8.8\%}              & {\color[HTML]{036400} 70.7\%}             \\
\multirow{-6}{*}{PV-A}              & \multirow{-2}{*}{Lyft}            & Waymo                             & {\color[HTML]{FE0000} -0.7}  & {\color[HTML]{FE0000} -3.8}  & {\color[HTML]{FE0000} -1.4\%}             & {\color[HTML]{FE0000} -13.6\%}            \\ \hline
                                    &                                   & nuScenes                          & {\color[HTML]{036400} +3.3}  & {\color[HTML]{036400} +3.1}  & {\color[HTML]{036400} 21.8\%}             & {\color[HTML]{036400} 60.4\%}             \\
                                    & \multirow{-2}{*}{Waymo}           & Lyft                              & {\color[HTML]{036400} +1.4}  & {\color[HTML]{036400} +1.0}  & {\color[HTML]{036400} 2.8\%}              & {\color[HTML]{036400} 2.8\%}              \\ \cline{2-7} 
                                    &                                   & Lyft                              & {\color[HTML]{036400} +15.8} & {\color[HTML]{036400} +11.5} & {\color[HTML]{036400} 85.0\%}             & {\color[HTML]{036400} 137.1\%}            \\
                                    & \multirow{-2}{*}{nuScenes}        & Waymo                             & {\color[HTML]{036400} +23.0} & {\color[HTML]{036400} +5.2}  & {\color[HTML]{036400} 172.0\%}            & {\color[HTML]{036400} 332.5\%}            \\ \cline{2-7} 
                                    &                                   & nuScenes                          & {\color[HTML]{036400} +0.9}  & {\color[HTML]{036400} +2.2}  & {\color[HTML]{036400} 6.4\%}              & {\color[HTML]{036400} 94.6\%}             \\
\multirow{-6}{*}{PV-C}              & \multirow{-2}{*}{Lyft}            & Waymo                             & {\color[HTML]{FE0000} -0.7}  & {\color[HTML]{FE0000} -6.2}  & {\color[HTML]{FE0000} -1.5\%}             & {\color[HTML]{FE0000} -23.4\%}            \\ \hline
                                    &                                   & nuScenes                          & {\color[HTML]{036400} +2.4}  & {\color[HTML]{036400} +1.9}  & {\color[HTML]{036400} 17.2\%}             & {\color[HTML]{036400} 53.0\%}             \\
                                    & \multirow{-2}{*}{Waymo}           & Lyft                              & {\color[HTML]{036400} +1.9}  & {\color[HTML]{036400} +3.5}  & {\color[HTML]{036400} 4.0\%}              & {\color[HTML]{036400} 11.4\%}             \\ \cline{2-7} 
                                    &                                   & Lyft                              & {\color[HTML]{036400} +15.5} & {\color[HTML]{036400} +11.0} & {\color[HTML]{036400} 111.3\%}            & {\color[HTML]{036400} 183.3\%}            \\
                                    & \multirow{-2}{*}{nuScenes}        & Waymo                             & {\color[HTML]{036400} +25.0} & {\color[HTML]{036400} +7.3}  & {\color[HTML]{036400} 348.4\%}            & {\color[HTML]{036400} \textgreater 500\%} \\ \cline{2-7} 
                                    &                                   & nuScenes                          & {\color[HTML]{036400} +1.3}  & {\color[HTML]{036400} +2.1}  & {\color[HTML]{036400} 9.6\%}              & {\color[HTML]{036400} 63.8\%}             \\
\multirow{-6}{*}{VX-A}             & \multirow{-2}{*}{Lyft}            & Waymo                             & {\color[HTML]{036400} +1.2}  & {\color[HTML]{036400} +5.9}  & {\color[HTML]{036400} 2.5\%}              & {\color[HTML]{036400} 20.1\%}             \\ \hline
                                    &                                   & nuScenes                          & {\color[HTML]{036400} +2.2}  & {\color[HTML]{036400} +1.8}  & {\color[HTML]{036400} 14.4\%}             & {\color[HTML]{036400} 36.5\%}             \\
                                    & \multirow{-2}{*}{Waymo}           & Lyft                              & {\color[HTML]{036400} +0.0}  & {\color[HTML]{036400} +2.9}  & {\color[HTML]{036400} 0.1\%}              & {\color[HTML]{036400} 8.4\%}              \\ \cline{2-7} 
                                    &                                   & Lyft                              & {\color[HTML]{036400} +20.4} & {\color[HTML]{036400} +13.8} & {\color[HTML]{036400} 185.5\%}            & {\color[HTML]{036400} 236.5\%}            \\
                                    & \multirow{-2}{*}{nuScenes}        & Waymo                             & {\color[HTML]{036400} +30.5} & {\color[HTML]{036400} +11.5} & {\color[HTML]{036400} \textgreater 500\%} & {\color[HTML]{036400} \textgreater 500\%} \\ \cline{2-7} 
                                    &                                   & nuScenes                          & {\color[HTML]{036400} +1.0}  & {\color[HTML]{036400} +2.2}  & {\color[HTML]{036400} 7.9\%}              & {\color[HTML]{036400} 90.0\%}             \\
\multirow{-6}{*}{VX-C}             & \multirow{-2}{*}{Lyft}            & Waymo                             & {\color[HTML]{036400} +1.7}  & {\color[HTML]{036400} +7.0}  & {\color[HTML]{036400} 3.7\%}              & {\color[HTML]{036400} 22.3\%}             \\ \hline
                                    &                                   & nuScenes                          & {\color[HTML]{036400} +1.1}  & {\color[HTML]{FE0000} -0.2}  & {\color[HTML]{036400} 10.0\%}             & {\color[HTML]{FE0000} -4.6\%}             \\
                                    & \multirow{-2}{*}{Waymo}           & Lyft                              & {\color[HTML]{036400} +0.3}  & {\color[HTML]{FE0000} -7.5}  & {\color[HTML]{036400} 0.6\%}              & {\color[HTML]{FE0000} -24.0\%}            \\ \cline{2-7} 
                                    &                                   & Lyft                              & {\color[HTML]{FE0000} -2.9}  & {\color[HTML]{FE0000} -1.3}  & {\color[HTML]{FE0000} -10.6\%}            & {\color[HTML]{FE0000} -13.3\%}            \\
                                    & \multirow{-2}{*}{nuScenes}        & Waymo                             & {\color[HTML]{036400} +5.0}  & {\color[HTML]{036400} +0.2}  & {\color[HTML]{036400} 23.0\%}             & {\color[HTML]{036400} 6.3\%}              \\ \cline{2-7} 
                                    &                                   & nuScenes                          & {\color[HTML]{FE0000} -0.7}  & {\color[HTML]{036400} +0.9}  & {\color[HTML]{FE0000} -8.3\%}             & {\color[HTML]{036400} 41.7\%}             \\
\multirow{-6}{*}{IA-SSD}            & \multirow{-2}{*}{Lyft}            & Waymo                             & {\color[HTML]{FE0000} -1.9}  & {\color[HTML]{FE0000} -9.3}  & {\color[HTML]{FE0000} -5.5\%}             & {\color[HTML]{FE0000} -52.7\%}            \\ \hline
\end{tabular}
\vspace{-4mm}
\end{table}

\subsection{Training with Accumulated Point Clouds}
\label{sec:train_with_accum}
First we explore the popular trend of point cloud accumulation in the cross-domain context, specifically for short sequence accumulation (i.e., less than 0.5s of historical frames). Directly concatenating the points from multiple historical frames has been demonstrated to improve performance for multiple 3D detectors \cite{yin2021centerpoint,shi2020pv,liu2023bevfusion,chen2022mppnet, yang20213dman} across various datasets \cite{caesar2020nuscenes, sun2020waymo} due to increased point density of objects. Accumulation is implemented by transforming past point clouds to the coordinate system of the current frame and appending a scalar timestamp $T$ indicating the time delta in seconds from the current frame. We denote the maximum time delta as $T_{\Delta \text{MAX}}$. For instance, for a 10Hz lidar where each point cloud is 0.1s apart, the accumulated point cloud at time $t$ can be represented as $P_t=\{P_{0}, P_{0.1}, P_{0.2},...,P_{N(0.1)}\}$ for $N$ historical frames with each point represented as $XYZT$ with $T_{\Delta \text{MAX}}=N(0.1)$. 

Currently, most UDA works for 3D object detection focus on single-frame detector adaptation \cite{yang2021st3d, you2022exploiting, tsai2022see, zhang2023uni3d,hu2023density}. However, in \cref{tab:analysis_multiframe} we demonstrate that simply training a detector on short sequence, accumulated point clouds can significantly improve domain generalization across multiple target domains for both vehicles and pedestrians. Furthermore, \cref{tab:multiframe_comparison_with_st3d} shows that multi-frame pre-training of SECOND-IoU on the Waymo dataset can even outperform ST3D \cite{yang2021st3d} on the nuScenes dataset without any domain adaptation. For training $\mathbf{D}_\text{multi}$, we use $T_{\Delta \text{MAX}}=0.45$ (10 frames) for nuScenes,  $T_{\Delta \text{MAX}}=0.3$ (4 frames) for Waymo and  $T_{\Delta \text{MAX}}=0.4$ (3 frames) for Lyft. For testing on different target domains in \cref{tab:analysis_multiframe}, we match the same $T_{\Delta \text{MAX}}$ as used in training. Empirically, the sweet spot for frame accumulation across datasets is around 0.4s which strikes a balance between dynamic and static objects, minimal motion tail overlap, and point cloud processing time. 

\setlength{\tabcolsep}{0.2em} 
\begin{table}[tb]
\centering
\caption{Pre-training SECOND-IoU with 4 frames on the Waymo dataset generally outperforms ST3D when testing on nuScenes data. Source and Target are given as Dataset / $T_{\Delta \text{MAX}}$.}
\label{tab:multiframe_comparison_with_st3d}
\begin{tabular}{c|c|c|cc|cc}
\hline
\multirow{2}{*}{Method} & \multirow{2}{*}{Source} & \multirow{2}{*}{Target} & \multicolumn{2}{c|}{Vehicle} & \multicolumn{2}{c}{Pedestrian} \\
                        &                         &                         & $\text{AP}_\text{BEV}$           & $\text{AP}_\text{3D}$           & $\text{AP}_\text{BEV}$            & $\text{AP}_\text{3D}$            \\ \hline
Source-Only             & W / 0.0                   & N / 0.0                   & 32.10         & 17.77        & 7.32           & 5.01          \\
ST3D \cite{yang2021st3d}                    & W / 0.0                   & N / 0.0                   & 35.90         & 20.20        & 5.75           & 5.11          \\
ST3D++ \cite{yang2022st3d++}                  & W / 0.0                   & N / 0.0                   & 35.73         & \textbf{20.90}        & 12.19          & 8.91          \\
Source-Only             & W / 0.3                   & N / 0.3                  & \textbf{40.36}         & 18.81        & \textbf{15.90}          & \textbf{9.05}          \\ \hline
\end{tabular}
\begin{tablenotes}\footnotesize
    \item [a] N: nuScenes, W: Waymo
\end{tablenotes} 
\vspace{-2mm}
\end{table}

\cref{tab:analysis_multiframe} indicates that training the same detector on accumulated point clouds performs significantly better on most cross-domain pairs compared to single-frame pre-training. In particular, multi-frame pre-training is highly beneficial for nuScenes-trained detectors, achieving up to $+30.5 \text{AP}_\text{3D}$ improvement when tested on Lyft/Waymo compared to single-frame pre-training. This is predominantly because accumulating sparse 32-beam point clouds can give variation in scan pattern representations of objects in the training dataset, especially with different ego-vehicle motion (straight roads/turning corners). This can include particular vehicle point representations that are common at different ranges in the Waymo or Lyft dataset, leading to an increased $\text{AP}_\text{3D}$ overall. Similarly, we show that multi-frame pre-training can also improve performance of Waymo detectors on Lyft and nuScenes point clouds. Notably, this improvement is not solely attributed to the non-rigid nature of vehicles, as pedestrians likewise achieve a significant boost in $\text{AP}_\text{3D}$ across most source-target pairs. 


$\mathbf{D}_\text{multi}$ however does not always improve cross-domain performance. The ability to leverage multi-frame point cloud training is detector-specific. Point-based IA-SSD shows mixed results when comparing $\mathbf{D}_\text{multi}$ and $\mathbf{D}_\text{single}$. This might be attributed to the learnable center-aware sampling being overfitted to the scan pattern of the training dataset, including how the accumulated points are distributed. There are also mixed results when feeding a Lyft-trained $\mathbf{\text{PV-RCNN}}_\text{multi}$ with Waymo data, potentially due to the ball-query sampling of accumulated historical points causing a discrepancy in object representations across domains. Due to the overwhelmingly positive improvement of $\mathbf{D}_\text{multi}$ compared to $\mathbf{D}_\text{single}$, we adopt multi-frame pre-training of all detectors for ensembling in MS3D++. Point cloud accumulation has the added benefit of greater point density at farther ranges, therefore increasing the box proposal recall at farther ranges as well.

\begin{table}[t]
\centering
\setlength{\tabcolsep}{0.4em} 
\caption{Varying the number of inference frames for multi-frame PV-C on cross-domain data. The source and target columns are written as dataset / $T_{\Delta \text{MAX}}$. Results reported with $\text{AP}_\text{3D}$ at IoU=0.7 and 0.5 for ``Vehicle" and ``Pedestrian" respectively.}
\label{tab:analysis_vmfi}
\begin{tabular}{c|c|ccc|ccc}
\hline
\multirow{2}{*}{Source} & \multirow{2}{*}{Target} & \multicolumn{3}{c|}{Vehicle}              & \multicolumn{3}{c}{Pedestrian}               \\ \cline{3-8} 
                                 &                                  & 0-30 & 30-50 & 50-80 & 0-30 & 30-50 & 50-80 \\ \hline
\multirow{7}{*}{N / 0.45}        & L / 0.0                          & 71.6          & \textbf{27.5}  & \textbf{3.9}   & 35.9          & 4.4            & 0.2            \\
                                 & L / 0.2                          & 71.9          & 25.7           & 3.4            & 43.3          & 10.3           & 1.6            \\
                                 & L / 0.4                          & \textbf{72.6} & 24.9           & 3.1            & \textbf{45.8} & \textbf{12.3}  & \textbf{1.8}   \\ \cline{2-8} 
                                 & W / 0.0                          & \textbf{74.6} & \textbf{29.5}  & \textbf{5.1}   & 14.0          & 4.4            & 0.4            \\
                                 & W / 0.1                          & 74.6          & 28.5           & 4.7            & 13.4          & 4.5            & 0.6            \\
                                 & W / 0.2                          & 74.3          & 28.1           & 4.6            & 14.2          & 4.2            & 0.6            \\
                                 & W / 0.4                          & 73.7          & 27.3           & 4.4            & \textbf{14.9} & \textbf{4.6}   & \textbf{0.7}   \\ \hline
\multirow{7}{*}{W / 0.3}         & L / 0.0                          & \textbf{80.3} & 55.0           & 15.4           & 51.7          & 23.3           & 7.1            \\
                                 & L / 0.2                          & 79.5          & \textbf{56.2}  & \textbf{17.6}  & \textbf{55.7} & \textbf{33.2}  & \textbf{20.7}  \\
                                 & L / 0.4                          & 79.3          & 55.1           & 17.1           & 51.5          & 27.7           & 18.4           \\ \cline{2-8} 
                                 & N / 0.1                          & 49.6          & 1.9            & 0.0            & 20.7          & 0.2            & 0.0            \\
                                 & N / 0.2                          & 51.3          & 2.7            & 0.1            & 23.3          & 0.4            & 0.0            \\
                                 & N / 0.3                          & \textbf{52.0} & 3.4            & 0.1            & \textbf{24.1} & 0.8            & 0.1            \\
                                 & N / 0.4                          & 51.7          & \textbf{3.7}   & \textbf{0.2}   & 22.4          & \textbf{1.1}   & \textbf{0.1}   \\ \hline
\end{tabular}
\begin{tablenotes}\footnotesize
    \item [a] N: nuScenes, W: Waymo, L: Lyft
\end{tablenotes} 
\vspace{-2mm}
\end{table}

\begin{table}[]
\centering
\setlength{\tabcolsep}{0.5em} 
\caption{Evaluation of 3D detectors from different sources on \textbf{Waymo} as the target domain. Results reported with $\text{AP}_\text{3D}$ at IoU=0.7 and 0.5 for ``Vehicle" and ``Pedestrian" respectively.}
\label{tab:analysis_target_waymo}
\begin{tabular}{l|c|ccc|ccc}
\hline
\multirow{2}{*}{Detector} & \multirow{2}{*}{Source} & \multicolumn{3}{c|}{Vehicle}              & \multicolumn{3}{c}{Pedestrian}               \\ \cline{3-8} 
                                 &                                  & 0-30 & 30-50 & 50-80 & 0-30 & 30-50 & 50-80 \\ \hline
PV-A                               & Waymo                            & 90.4          & 68.2           & 40.7           & 67.7          & 56.7           & 37.0           \\
PV-C                               & Waymo                            & 91.1          & 70.1           & 42.3           & 75.9          & 68.1           & 53.0           \\
VX-A                              & Waymo                            & 90.8          & 69.8           & 43.6           & 68.9          & 61.0           & 46.6           \\
VX-C                              & Waymo                            & \textbf{91.1} & \textbf{71.7}  & \textbf{45.5}  & \textbf{78.5} & \textbf{71.7}  & \textbf{60.3}  \\
IA-SSD                             & Waymo                            & 86.7          & 59.7           & 31.3           & 60.9          & 55.4           & 42.6           \\ \hline\hline
\multirow{2}{*}{PV-A}              & nuScenes                         & 73.7          & 28.9           & 4.7            & 12.2          & 3.5            & 0.5            \\
                                   & Lyft                             & 75.7          & 46.9           & 23.1           & 40.3          & 29.3           & 14.9           \\ \hline
\multirow{2}{*}{PV-C}              & nuScenes                         & 74.6          & 29.5           & 5.1            & 14.9          & 4.6            & 0.7            \\
                                   & Lyft                             & 73.9          & 43.7           & 21.4           & 35.2          & 27.7           & 16.6           \\ \hline
\multirow{2}{*}{VX-A}             & nuScenes                         & 71.2          & 21.5           & 3.9            & 21.5          & 2.7            & 0.5            \\
                                   & Lyft                             & \textbf{75.7} & \textbf{48.1}  & \textbf{24.8}  & 46.1          & 37.0           & 22.0           \\ \hline
\multirow{2}{*}{VX-C}             & nuScenes                         & 71.5          & 21.8           & 4.2            & 27.5          & 6.1            & 1.4            \\
                                   & Lyft                             & 74.0          & 46.8           & 24.5           & \textbf{46.8} & \textbf{40.8}  & \textbf{28.3}  \\ \hline
\multirow{2}{*}{IA-SSD}            & nuScenes                         & 52.1          & 22.5           & 4.9            & 5.8           & 4.2            & 1.3            \\
                                   & Lyft                             & 59.9          & 31.9           & 12.5           & 23.1          & 19.0           & 10.9           \\ \hline
\end{tabular}
\vspace{-2mm}
\end{table}

\subsection{Varying Multi-frame Inference}
\label{sec:vmfi}
When using multi-frame detectors for cross-domain data however, a pertinent question arises: Should we prioritize matching $T_{\Delta \text{MAX}}$ or point density for achieving optimal performance? For instance, accumulating 0.45 seconds of nuScenes data results in around 350k points, while for Waymo, 0.4s corresponds to 850k points, thereby further amplifying the scan pattern domain gap. In \cref{tab:analysis_vmfi}, we show that varying $T_{\Delta \text{MAX}}$ at inference can lead to different results across ranges and classes. When testing a nuScenes-trained PV-C on Lyft, the optimal performance for vehicles at 30-80m is without point cloud accumulation, whereas pedestrians gain up to $+10 \text{AP}_\text{3D}$ from $T_{\Delta \text{MAX}}=0.4$ compared to no accumulation. Overall, it is challenging to select the optimal $T_{\Delta \text{MAX}}$ as it is difficult to predict how a detector trained on one scan pattern will generalize to another scan pattern. In MS3D++, we take advantage of this insight and generate multiple bounding box sets $\{B^i_{m}\}^M_{m=0}$ by varying $T_{\Delta \text{MAX}}$ at inference to consider all possible scan patterns. Each varied $T_{\Delta \text{MAX}}$ at inference is another set of bounding boxes $B^i_m$ for the $m$-th variation on the $i$-th frame for a total of $M$ sets of bounding boxes. We demonstrate in \cref{sec:ablation} that we can substantially increase both precision and recall with this approach.

\begin{table}[]
\centering
\setlength{\tabcolsep}{0.5em} 
\caption{Evaluation of 3D detectors from different sources on \textbf{nuScenes} as the target domain. Results reported with $\text{AP}_\text{3D}$ at IoU=0.7 and 0.5 for ``Vehicle" and ``Pedestrian" respectively.}
\label{tab:analysis_target_nuscenes}
\begin{tabular}{l|c|ccc|ccc}
\hline
\multirow{2}{*}{Detector} & \multirow{2}{*}{Source} & \multicolumn{3}{c|}{Vehicle}              & \multicolumn{3}{c}{Pedestrian}               \\ \cline{3-8} 
                                 &                                  & 0-30 & 30-50 & 50-80 & 0-30 & 30-50 & 50-80 \\ \hline
PV-A                               & nuScenes                         & \textbf{72.6} & \textbf{20.8}  & \textbf{2.6}   & \textbf{44.0} & 13.8           & 1.4            \\
PV-C                               & nuScenes                         & 68.9          & 18.9           & 2.2            & 42.2          & \textbf{14.8}  & 1.4            \\
VX-A                              & nuScenes                         & 69.8          & 17.2           & 2.1            & 42.7          & 12.3           & 0.9            \\
VX-C                              & nuScenes                         & 66.6          & 17.5           & 1.9            & 43.2          & 14.8           & \textbf{1.7}   \\
IA-SSD                             & nuScenes                         & 57.0          & 10.2           & 0.8            & 31.5          & 8.9            & 0.7            \\ \hline \hline
\multirow{2}{*}{PV-A}              & Waymo                            & 51.0          & 3.2            & 0.1            & 22.2          & 0.8            & 0.0            \\
                                   & Lyft                             & 44.9          & 2.5            & 0.1            & 19.1          & 0.8            & 0.0            \\ \hline
\multirow{2}{*}{PV-C}              & Waymo                            & \textbf{52.0} & \textbf{3.4}   & \textbf{0.1}   & \textbf{24.1} & \textbf{0.8}   & \textbf{0.1}   \\
                                   & Lyft                             & 40.8          & 2.0            & 0.0            & 13.4          & 0.3            & 0.1            \\ \hline
\multirow{2}{*}{VX-A}             & Waymo                            & 46.1          & 2.6            & 0.0            & 16.4          & 0.3            & 0.0            \\
                                   & Lyft                             & 42.6          & 1.8            & 0.1            & 16.0          & 0.3            & 0.0            \\ \hline
\multirow{2}{*}{VX-C}             & Waymo                            & 48.6          & 2.8            & 0.1            & 19.0          & 0.8            & 0.0            \\
                                   & Lyft                             & 41.3          & 1.7            & 0.1            & 14.0          & 0.2            & 0.0            \\ \hline
\multirow{2}{*}{IA-SSD}            & Waymo                            & 35.0          & 0.7            & 0.0            & 12.2          & 0.1            & 0.0            \\
                                   & Lyft                             & 23.0          & 0.8            & 0.0            & 8.5           & 0.5            & 0.0            \\ \hline
\end{tabular}
\end{table}

\begin{table}[]
\centering
\setlength{\tabcolsep}{0.5em} 
\caption{Evaluation of 3D detectors from different sources on \textbf{Lyft} as the target domain. Results reported with $\text{AP}_\text{3D}$ at IoU=0.7 and 0.5 for ``Vehicle" and ``Pedestrian" respectively.}
\label{tab:analysis_target_lyft}
\begin{tabular}{l|c|ccc|ccc}
\hline
\multirow{2}{*}{Detector} & \multirow{2}{*}{Source} & \multicolumn{3}{c|}{Vehicle}              & \multicolumn{3}{c}{Pedestrian}               \\ \cline{3-8} 
                                 &                                  & 0-30 & 30-50 & 50-80 & 0-30 & 30-50 & 50-80 \\ \hline
PV-A                               & Lyft                             & \textbf{90.3} & \textbf{73.3}  & \textbf{29.0}  & 57.0          & 33.4           & \textbf{19.2}  \\
PV-C                               & Lyft                             & 88.2          & 71.1           & 27.8           & 53.9          & 33.1           & 17.5           \\
VX-A                              & Lyft                             & 88.0          & 70.7           & 26.2           & 57.8          & \textbf{36.6}  & 18.9           \\
VX-C                              & Lyft                             & 88.0          & 70.4           & 26.3           & \textbf{59.5} & 34.4           & 18.9           \\
IA-SSD                             & Lyft                             & 82.6          & 58.7           & 17.6           & 28.9          & 18.9           & 12.2           \\ \hline\hline
\multirow{2}{*}{PV-A}              & nuScenes                         & 73.1          & 27.7           & 3.8            & 39.1          & 11.0           & 1.4            \\
                                   & Waymo                            & 78.8          & 55.8           & 17.2           & 57.1          & \textbf{34.0}  & 18.2           \\ \hline
\multirow{2}{*}{PV-C}              & nuScenes                         & 71.6          & 27.5           & 3.9            & 45.8          & 12.3           & 1.8            \\
                                   & Waymo                            & \textbf{79.5} & \textbf{56.2}  & \textbf{17.6}  & 55.7          & 33.2           & \textbf{20.7}  \\ \hline
\multirow{2}{*}{VX-A}             & nuScenes                         & 68.0          & 18.3           & 2.2            & 39.6          & 9.9            & 1.6            \\
                                   & Waymo                            & 78.0          & 53.9           & 16.5           & 55.4          & 30.9           & 16.2           \\ \hline
\multirow{2}{*}{VX-C}             & nuScenes                         & 69.9          & 22.1           & 2.4            & 45.5          & 11.9           & 1.6            \\
                                   & Waymo                            & 75.9          & 54.1           & 16.0           & \textbf{57.8} & 33.6           & 18.6           \\ \hline
\multirow{2}{*}{IA-SSD}            & nuScenes                         & 52.0          & 19.4           & 3.1            & 17.3          & 6.2            & 2.0            \\
                                   & Waymo                            & 74.8          & 48.9           & 12.0           & 54.3          & 28.7           & 11.1           \\ \hline
\end{tabular}
\vspace{-2mm}
\end{table}

\subsection{Cross-Domain Results for Each Target Domain}
\label{sec:study_cross_domain_results}
\cref{tab:analysis_target_waymo,tab:analysis_target_lyft,tab:analysis_target_nuscenes} presents a comprehensive assessment of the overall performance of each detector in a cross-domain setting, evaluated on Waymo, Lyft and nuScenes datasets. We observe that 3D detectors perform reasonably well on close range (0-30m) detection across various target domains, irrespective of their source domain. This can be attributed to the fact that beam divergence across different lidars is less prominent at closer ranges, therefore having less of a domain shift. However, as the range increases, the scan pattern domain gap is exacerbated due to differences in beam divergence, leading to drastic performance degradation in detectors. Interestingly, no single detector or detection head stands out as the highest performer, indicating that existing state-of-the-art 3D detectors are far from domain generalization on point cloud data. However, we observe that voxel-based backbone networks (PV-RCNN++ and VoxelRCNN) perform better than IA-SSD on all target domains, potentially due to the limitations of IA-SSD's learnable sampling module in domain generalization. In \cref{tab:analysis_ap_reduction}, we breakdown the degradation of each detector when adapting from different source domains to a given target domain, compared to training from scratch on that target domain. From this table, we find that the drop in detector performance is more impacted by the source-target domain pairing rather than the detector architecture itself. In general, source-target pairs that are more similar to each other such as Lyft and Waymo, which both contain 64-beam point cloud data have less of a performance drop compared to Waymo/Lyft to nuScenes.

\begin{table}[]
\centering
\setlength{\tabcolsep}{0.5em} 
\caption{Performance reduction when comparing pre-trained detectors on different source domains $\text{AP}_\text{CD}$ to training from scratch on target domain data $\text{AP}_\text{oracle}$ where $\text{AP}_\text{\% Reduction} = \frac{\text{AP}_\text{CD}-\text{AP}_\text{oracle}}{\text{AP}_\text{oracle}} \times 100$.}
\label{tab:analysis_ap_reduction}
\begin{tabular}{l|l|l|cc}
\hline
                           &                            &                            & \multicolumn{2}{c}{\% $\text{AP}_{\text{3D}}$ Reduction}                               \\ \cline{4-5} 
\multirow{-2}{*}{Source}   & \multirow{-2}{*}{Target}   & \multirow{-2}{*}{Detector} & Veh.                       & Ped.                            \\ \hline
                           &                            & PV-A                       & \cellcolor[HTML]{FF6161}-69.9\% & \cellcolor[HTML]{FF3A3A}-80.8\% \\
                           &                            & PV-C                       & \cellcolor[HTML]{FF6161}-66.8\% & \cellcolor[HTML]{FF4D4D}-78.0\% \\
                           &                            & VX-A                      & \cellcolor[HTML]{FF4D4D}-72.4\% & \cellcolor[HTML]{FF3A3A}-86.2\% \\
                           &                            & VX-C                      & \cellcolor[HTML]{FF6161}-69.2\% & \cellcolor[HTML]{FF3A3A}-82.9\% \\
\multirow{-5}{*}{Waymo}    & \multirow{-5}{*}{nuScenes} & IA-SSD                     & \cellcolor[HTML]{FF4D4D}-77.3\% & \cellcolor[HTML]{FF3A3A}-86.6\% \\ \hline
                           &                            & PV-A                       & \cellcolor[HTML]{FF4D4D}-74.3\% & \cellcolor[HTML]{FF3A3A}-83.1\% \\
                           &                            & PV-C                       & \cellcolor[HTML]{FF4D4D}-76.1\% & \cellcolor[HTML]{FF3A3A}-87.5\% \\
                           &                            & VX-A                      & \cellcolor[HTML]{FF4D4D}-74.5\% & \cellcolor[HTML]{FF3A3A}-86.4\% \\
                           &                            & VX-C                      & \cellcolor[HTML]{FF4D4D}-75.3\% & \cellcolor[HTML]{FE0000}-88.8\% \\
\multirow{-5}{*}{Lyft}     & \multirow{-5}{*}{nuScenes} & IA-SSD                     & \cellcolor[HTML]{FF3A3A}-83.1\% & \cellcolor[HTML]{FE0000}-88.9\% \\ \hline
                           &                            & PV-A                       & \cellcolor[HTML]{FF9C9C}-30.3\% & \cellcolor[HTML]{FF8888}-49.5\% \\
                           &                            & PV-C                       & \cellcolor[HTML]{FF9C9C}-35.3\% & \cellcolor[HTML]{FF6161}-60.6\% \\
                           &                            & VX-A                      & \cellcolor[HTML]{FF9C9C}-30.3\% & \cellcolor[HTML]{FF8888}-41.7\% \\
                           &                            & VX-C                      & \cellcolor[HTML]{FF9C9C}-33.3\% & \cellcolor[HTML]{FF8888}-45.5\% \\
\multirow{-5}{*}{Lyft}     & \multirow{-5}{*}{Waymo}    & IA-SSD                     & \cellcolor[HTML]{FF8888}-45.8\% & \cellcolor[HTML]{FF6161}-67.4\% \\ \hline
                           &                            & PV-A                       & \cellcolor[HTML]{FF7575}-54.9\% & \cellcolor[HTML]{FE0000}-91.5\% \\
                           &                            & PV-C                       & \cellcolor[HTML]{FF7575}-54.6\% & \cellcolor[HTML]{FE0000}-90.8\% \\
                           &                            & VX-A                      & \cellcolor[HTML]{FF6161}-60.6\% & \cellcolor[HTML]{FF3A3A}-87.8\% \\
                           &                            & VX-C                      & \cellcolor[HTML]{FF6161}-60.7\% & \cellcolor[HTML]{FF3A3A}-84.7\% \\
\multirow{-5}{*}{nuScenes} & \multirow{-5}{*}{Waymo}    & IA-SSD                     & \cellcolor[HTML]{FF6161}-62.1\% & \cellcolor[HTML]{FE0000}-93.3\% \\ \hline
                           &                            & PV-A                       & \cellcolor[HTML]{FF7575}-56.1\% & \cellcolor[HTML]{FF6161}-63.6\% \\
                           &                            & PV-C                       & \cellcolor[HTML]{FF7575}-55.4\% & \cellcolor[HTML]{FF7575}-55.9\% \\
                           &                            & VX-A                      & \cellcolor[HTML]{FF6161}-62.8\% & \cellcolor[HTML]{FF6161}-65.4\% \\
                           &                            & VX-C                      & \cellcolor[HTML]{FF6161}-60.1\% & \cellcolor[HTML]{FF6161}-60.2\% \\
\multirow{-5}{*}{nuScenes} & \multirow{-5}{*}{Lyft}     & IA-SSD                     & \cellcolor[HTML]{FF6161}-62.1\% & \cellcolor[HTML]{FF6161}-63.6\% \\ \hline
                           &                            & PV-A                       & \cellcolor[HTML]{FFB0B0}-25.7\% & \cellcolor[HTML]{FFC3C3}-1.0\%  \\
                           &                            & PV-C                       & \cellcolor[HTML]{FFB0B0}-22.5\% & \cellcolor[HTML]{CCEAD4}7.2\%   \\
                           &                            & VX-A                      & \cellcolor[HTML]{FFB0B0}-24.1\% & \cellcolor[HTML]{FFC3C3}-11.4\% \\
                           &                            & VX-C                      & \cellcolor[HTML]{FFB0B0}-25.3\% & \cellcolor[HTML]{FFC3C3}-2.2\%  \\
\multirow{-5}{*}{Waymo}    & \multirow{-5}{*}{Lyft}     & IA-SSD                     & \cellcolor[HTML]{FFC3C3}-19.3\% & \cellcolor[HTML]{6AC081}43.4\%  \\ \hline
\end{tabular}
\vspace{-2mm}
\end{table}

To delve deeper into the potential reasons for the significant performance decrease, we visualize the predictions and present a summary of several key findings in \cref{fig:dataset_domain_gap}. Despite dense observations of objects, pre-trained detectors may struggle to obtain detections, indicating that higher point density alone is insufficient for effective cross-domain detection. We observe that a nuScenes-trained detector misses many densely observed vehicles in \cref{fig:dataset_domain_gap} (2) potentially due to weather artefacts exhibiting different point distributions across lidar types. For \cref{fig:dataset_domain_gap} (2), it could also be attributed to the peculiar labeling guidelines of Lyft where labels do not often include non-traffic participants, therefore causing a Lyft-trained detector to have many missed detections of parked cars on other datasets. Another prominent issue arises when adapting from dense to sparse lidars. In the nuScenes dataset, the 32-beam lidar's large vertical angular resolution results in sparse information for vehicles beyond 30m, making accurate height localization challenging shown in \cref{fig:dataset_domain_gap}. While many existing works focus on the number of beams, we argue that the scan pattern domain gap is more related to the angular resolution. A 32-beam lidar with small vertical angular resolution can also result in dense vehicle appearances that are similar the 64-beam lidar in Waymo, albeit with a smaller vertical field of view. (3). Additionally, the Lyft dataset does not appear to perform ego-motion compensation, resulting in misaligned lidar points during a single revolution due to vehicle ego-motion\cite{berrio2022camera}. This introduces challenges for accurate detection and labeling consistency. As depicted in \cref{fig:dataset_domain_gap} (4), there are instances where the ground truth box includes the latest points in the point cloud, while in other cases, it disregards the latest points. In general, these various domain idiosyncrasies substantially affects predicted confidence scores. Many 3D detector works place a large importance on IoU-aware refinement to enable the confidence score to reflect the accuracy of box localization \cite{hu2022afdetv2,yang2021st3d,shi2020pv, yin2021centerpoint,jiang2018acquisition} because this scoring itself has a large impact on the AP metric. However, in the cross-domain setting, this relationship between confidence score and object localization becomes misaligned and reliable, therefore further affecting $\text{AP}_\text{3D}$. 

Another prominent aspect of the domain gap is due to labeling discrepancies. For example, Waymo labeling guidelines specify that side mirrors of vehicles are to be included in the label whereas nuScenes, Lyft and KITTI do not include side mirrors. This discrepancy can considerably affect the IoU calculation when comparing predictions with ground-truth boxes for detector evaluation. Additionally, there are inconsistent category definitions. The ``Vehicle" class of the Waymo dataset includes cars, trucks, buses, construction vehicles, and even motorcycles whereas nuScenes and Lyft classify them separately. This is an issue when adapting Waymo detectors to KITTI for example, as KITTI lacks specific categories for motorcycles and construction vehicles, leading to numerous ``false positive" detections when using a detector trained on the ``Vehicle" category from Waymo. Furthermore, there are labeling inconsistencies within the ``Cyclist" category across datasets. Waymo labels ``Cyclists" only if there is a rider, whereas Lyft and nuScenes have no ``Cyclist" category but rather an object-based, ``bicycle" category that includes parked bicycles without a rider. 

\section{MS3D++}
\label{sec:ms3d++}
\subsection{Overview}
The goal of MS3D++ is to use multiple pre-trained detectors from various source domains to automatically label an unlabeled target domain. We use multiple pre-trained detectors with their predictions denoted as the set of bounding boxes $\{B^i_{m}\}^M_{m=0}$ representing the $m$-th prediction set on the $i$-th frame of the unlabeled target dataset for a total of $M$ prediction sets. Each test-time augmentation or VMFI (\cref{sec:vmfi}) are also included in the $m$-th bounding box set. We consider a sequence of target domain point clouds $\{P_i \in \mathbf{R}^{n_i \times C}\}$, $i=1,2,...,N$ with the $i$-th frame point cloud $P_i$ consisting of $n_i$ points with $C$ channels for each point, for $N$ total frames. For the cross-domain setting, we only consider the channels $XYZT$ where $T$ is the time delta in seconds from the current frame at time $t=0$. For MS3D++ we assume known sensor poses $\{E_i = [R_i | t_i] \in \mathbf{R}^{3 \times 4}\}, i=1,2,...,N$ at each frame in world-coordinates for ego-motion compensation. 

MS3D++ is designed to achieve two primary goals in self-training: First, maximise precision through accurately localized and categorized labels to avoid reinforcing false positives. Secondly, we aim to maximise recall to ensure sufficiently accurately labeled samples for improving model performance without compromising precision. MS3D++ can be broken down into three key steps: (1) Increasing box proposal recall through ensembling and tracking; (2) Temporal refinement to improve precision; (3) Multi-stage training to boost recall while maintaining precision. \cref{fig:ms3d++_framework} illustrates our proposed MS3D++ framework which will be introduced in the following sub-sections.

\subsection{Kernel Density Estimation Box Fusion (KBF)}
In our previous work \cite{tsai2023ms3d}, we introduced KBF, $\kappa(\cdot)$, for effective fusion of multiple predicted boxes which uses Kernel Density Estimation (KDE) for fusing the centroid $(c_x,c_y,c_z)$, dimensions $(l,w,h)$, heading $\theta$ and a confidence score $s$ of a set of boxes separately.  KDE positions a kernel $K$ at each data point $x_i$ and sums them together to get a probability density estimate $\hat{f}(x)$, shown in \cref{eqn:kde}. A weight $\text{w}_i$ for each data point and a kernel bandwidth $h$ can be adjusted for fine-tuning the PDF estimate. KBF remains a core component for all components of MS3D++ with the additional use of $\text{w}_i$ for weighing of bounding box sets in \cref{sec:detector_weighing} using insights from \cref{sec:quantifying_cross_domain_detection}.

\begin{equation}
\hat{f}(x) = \frac{1}{h}\sum^{N}_{i=1}\text{w}_iK(\frac{x-x_i}{h})
\label{eqn:kde}
\end{equation}

\subsection{Increasing Box Proposal Recall}
\label{sec:increasing_box_proposal_recall}
A primary challenge for self-training in unsupervised domain adaptation is the lack of strong pre-trained detectors for pseudo-label generation. For example, as shown in \cref{sec:quantifying_cross_domain_detection}, training PV-RCNN++\cite{shi2023pv++} and VoxelRCNN\cite{deng2020voxelrcnn} on Waymo and testing on nuScenes data can result in up to 70\% decrease in Vehicle detection rate compared to training the same models from scratch with nuScenes data. This weak baseline can lead to a large number of missed detections and misaligned confidence scores. To achieve effective pseudo-labeling, it becomes crucial to maximise recall by increasing the number of object proposals, thereby increasing our ability to capture all objects within the scene (i.e., reducing false negatives). In 3D detector methods, it is also common to output a high number of predictions before culling down with NMS \cite{yin2021centerpoint} or a second-stage refinement \cite{chen2022mppnet,shi2020pv}. Following this, MS3D++ aims to first increase box proposal recall with detector ensembling and tracking for inter/extrapolation.

\subsubsection{Varied Multi-Frame Inference} 
\begin{figure}[t]
  \centering
  \includegraphics[width=0.99\linewidth]{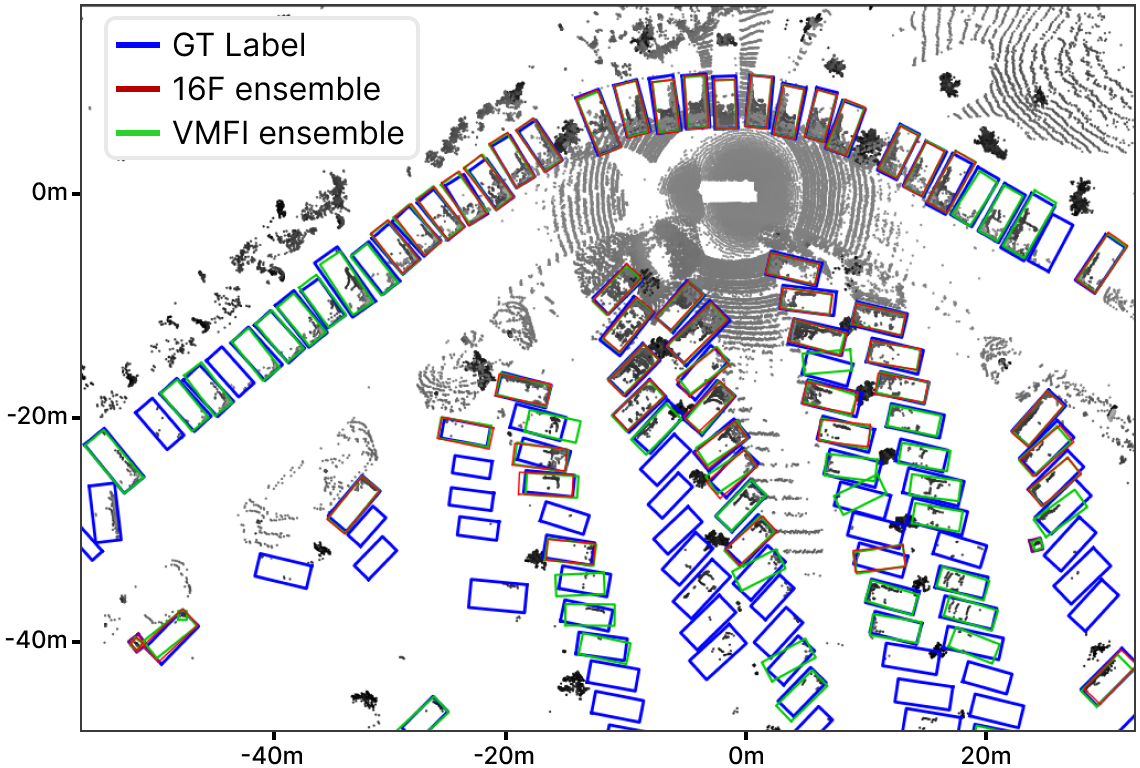}
  \caption{Ensembling with Varied Multi-frame Inference (VMFI) obtains an extended range of detections compared to using 16-frames for inference. This could be attributed to the exacerbated scan pattern domain gap when feeding 16-frames to a pre-trained multi-frame detector of a different source domain.}
  \label{fig:vmfi_vs_16f_ens}
  \vspace{-2mm}
\end{figure}

In \cref{fig:dataset_domain_gap}, we illustrated that certain scan pattern differences between datasets can cause pre-trained detectors to incorrectly predict a high number of false negatives (i.e., missed detections). To boost the recall in our initial box proposals, we first use the insights from \cref{tab:analysis_multiframe} and adopt multi-frame pre-trained detectors to raise the baseline performance for unseen target domains. Thereafter, following the findings of \cref{tab:analysis_vmfi}, we employ Varied Multi-Frame Inference (VMFI) which varies the number of accumulated frames at inference. With VMFI, we take the guess-work out of matching scan patterns, and simply include all variations in the ensemble. In contrast to our previous work \cite{tsai2023ms3d}, we remove the need for 16-frame detection sets as we show in \cref{fig:vmfi_vs_16f_ens} that VMFI can obtain detections for a much farther range of objects. Finally, we also generate detections for each VMFI with Test-Time Augmentation (TTA) for each detector. From these detection sets, we use KBF to obtain a single set of box proposals for each $i$-th frame $B^i_\text{KBF} = \kappa({\{B^i_{m\text{(VMFI)}},B^i_{m\text{(TTA)}}\}})$.

\subsubsection{Detector Weighing} \label{sec:detector_weighing} From \cref{sec:study_cross_domain_results}, we showed that each detector has varying performance on different target domains. However due to the idiosyncasies of particular source-target domain pairs, it is challenging to build an intuition of the optimal detector for specific target domains. Therefore to alleviate the strain on having to select optimal detectors, we simply include all the detectors into our ensemble with the exception of IA-SSD due to its low cross-domain performance. We show in \cref{sec:ablation} that KBF is able to successfully fuse the different detectors with minimal loss of performance. Furthermore, we devise a weighing strategy guided by simple intuition: assigning higher weight to detectors that were pre-trained on the lidar density most similar to the target domain. We avoid weighing individual detectors above another based on our observation in \cref{tab:analysis_ap_reduction}, where the performance of detectors on a new target domain is primarily influenced by its source dataset rather than its specific architecture. This approach also enables visual assessment of the ensembling performance, eliminating the need for a manually labeled validation set to determine the optimal combinations and weights.

\subsubsection{Multi-Object Tracking} Due to the misaligned confidence scores of pre-trained detectors in a cross-domain setting, there are often highly accurate detections with scores as low as 0.3. For example, with Waymo as the target domain, many accurate pedestrian predictions from nuScenes/Lyft detectors have scores in the range $[0.3,0.6]$. The challenge however, is separating accurate pedestrian boxes from the myriad of false positive detections that also exist within that range which we further explain in \cref{sec:temporal_refinement}. For this stage, to prevent pre-mature elimination of true positives, we run a Kalman Filter-based tracker \cite{pang2021simpletrack} at low score thresholds for pedestrians ($T_{\text{p}}$), and vehicles ($T_{\text{v,all}}$ and $T_{\text{v,static}}$) on $B_\text{KBF}$. We generate two separate tracks for vehicles at different thresholds: (1) $T_{\text{v,all}}$ tracks both dynamic and static vehicles, and (2) $T_{\text{v,static}}$ focuses on static vehicle tracking with a high BEV IoU threshold for track association. Following our prior work, we classify all tracks into static or dynamic based on pre-defined thresholds of begin-to-end distance and box centre variance. A high IoU for $T_{\text{v,static}}$ is critical in minimising false positive tracks of parked vehicles that are classified as ``dynamic" due to inaccurate box localization. While the tracking step includes both dynamic and static vehicles, extrapolation of static vehicles is simpler and less prone to false positives. For instance, when a tracked dynamic vehicle turns a corner, becomes occluded, or changes lanes at a far range, the extrapolated track can easily deviate significantly from the actual path or suffer from poor box localization. To address this, we use interpolation and extrapolations only for confident dynamic tracks by setting a higher score threshold for $T_{\text{v,all}}$ than $T_{\text{v,static}}$.

\subsection{Temporal Refinement}
\label{sec:temporal_refinement}
Due to misaligned confidence scores in a cross-domain setting, it is often challenging to decide if a predicted box is a true or a false positive for filtering of box proposals. Simply setting a confidence score threshold often proves inadequate as high confidence scores may not correlate with an accurate box localization or object class. Furthermore, selecting a threshold involves balancing between data variation and quality. A high threshold may reduce false positives, but limit the data variation and as a result, model improvement. Conversely, a low threshold may increase data variation but lead to more false positives and poorly localized boxes, resulting in the accumulation of errors with each self-training round. In this section, we explain how MS3D++ leverages temporal consistency and object characteristics to provide a more effective solution for refining box proposals to obtain increased pseudo-label precision. For the final pseudo-label set, we define a positive score threshold $s_{\text{pos}}$ such that every box with a score greater than $s_{\text{pos}}$ is considered a pseudo-label. 

\subsubsection{Retroactive object labeling} \label{sec:track_refinement} In human perception, identifying certain objects from a distance or when occluded can be challenging. To address this, our eyes continuously track the object until its class can be confidently determined. Once recognized, we retroactively apply this object class label to all past and future observations of the same object. In the context of MS3D++, we adopt a similar approach, where we track detections with a low threshold to obtain potential class proposals. These tracks are not immediately treated as pseudo-labels. We withhold the use of the tracks as labels until we confidently recognize the object class, through a number of high confidence detections (i.e., \textgreater \space $s_{\text{pos}}$). Practically, since pseudo-labeling is done in an offline setting, we simply filter every track to ensure they contain at least $N_\text{pos}$ confident detections above score $s_{\text{pos}}$ to obtain the refined tracks: $T_{\text{v,all(refined)}}$,$T_{\text{v,static(refined)}}$ and $T_{\text{p(refined)}}$. We note that this approach is mostly effective for $T_{\text{v,static}}$ and $T_{\text{p}}$ which further elaborate on.

\subsubsection{Static vehicle refinement} For dynamic vehicle tracking, although interpolated or extrapolated tracks may provide good centre prediction, the tracked box is impractical as a pseudo-label if its orientation is incorrect. Due to the elongated shape of vehicles, heading errors can easily exclude most of the vehicle, therefore resulting in poorly localized pseudo-labels that affect the self-training process. This particularly affects the extrapolation of sparse vehicle observations at farther ranges. Stationary vehicles on the other hand are much easier to handle. For $T_{\text{v,static}}$, we simply require a few strong detections of the object to be able to use it for refining all tracked observations of the stationary vehicle. Therefore, to increase the precision of vehicle labeling, we focus on scenes with static vehicles. Additionally, because vehicles are rigid body objects, dynamic vehicles share the same appearance as static vehicles. This allows detectors trained on accurately labeled static vehicles to effectively handle dynamic vehicles as well. Following our previous work, we adopt KBF for fusing the box predictions of $H$ historical observation frames, $B^{i,j}_\text{v,static} = \kappa(\{T_{\text{v,static}}\}^i_{i-H})$, to obtain a static box $B^{i,j}_\text{v,static}$ for the $j$-th vehicle in the $i$-th frame due to potential localization errors \cite{tsai2023ms3d,caesar2020nuscenes}. 

\subsubsection{Pedestrian refinement} Pre-trained detectors commonly misclassify pole-like objects as ``Pedestrians". The confidence scores of such predictions fall within a similar range (around 0.3-0.6) to actual ``Pedestrians" in the scene. Since pedestrians are non-rigid body objects that are frequently in motion, the static refinement approach used for ``Vehicles" is not directly applicable. However, we leverage this unique characteristic of pedestrians to our advantage. While pole-like objects and pedestrians may appear similar when static, the distinction becomes clearer when the object starts moving. Therefore, we classify the motion of tracks from $T_{\text{p(refined)}}$ and use the dynamic tracks $T_{\text{p(refined),dyn}}$ as pseudo-labels. This strategy is more practical than dynamic vehicle tracking due to the geometric property of pedestrians. Unlike vehicles, the approximate symmetry of pedestrians from a BEV perspective allows for relatively high IoU with ground truth even with incorrect heading predictions. As a result, obtaining reasonably localized pedestrian pseudo-labels with retroactive object labeling on $T_{\text{p(refined),dyn}}$ is considerably more feasible compared to dynamic vehicles. Nonetheless, pedestrian labeling need not solely be limited to dynamic pedestrians. After a few rounds of self-training, explained in \cref{sec:multi_stage_self_training}, detector confidence scores become better calibrated, making it easy to distinguish true pedestrians by thresholding. At this point, retroactive object labeling can be used for both static and dynamic pedestrians.

\subsection{Pseudo-Label Filtering}
To obtain our final set of pseudo-labels, we add the following boxes to each $i$-th frame and apply NMS for filtering: (1) VMFI Boxes $B^i_\text{KBF}$, (2) Static vehicle boxes $B^i_\text{v,static}$, (3) Tracked vehicle boxes $T^i_\text{v,all(refined)}$, (4) Dynamic pedestrian tracked boxes $T^i_{\text{p(refined),dyn}}$. Following this, we remove any pseudo-labels with less than 1 point in the box for a single-frame point cloud and only retain pseudo-labels above a score threshold $s \geq s_\text{pos}$.

\begin{figure}[t]
  \centering
  \includegraphics[width=0.9\linewidth]{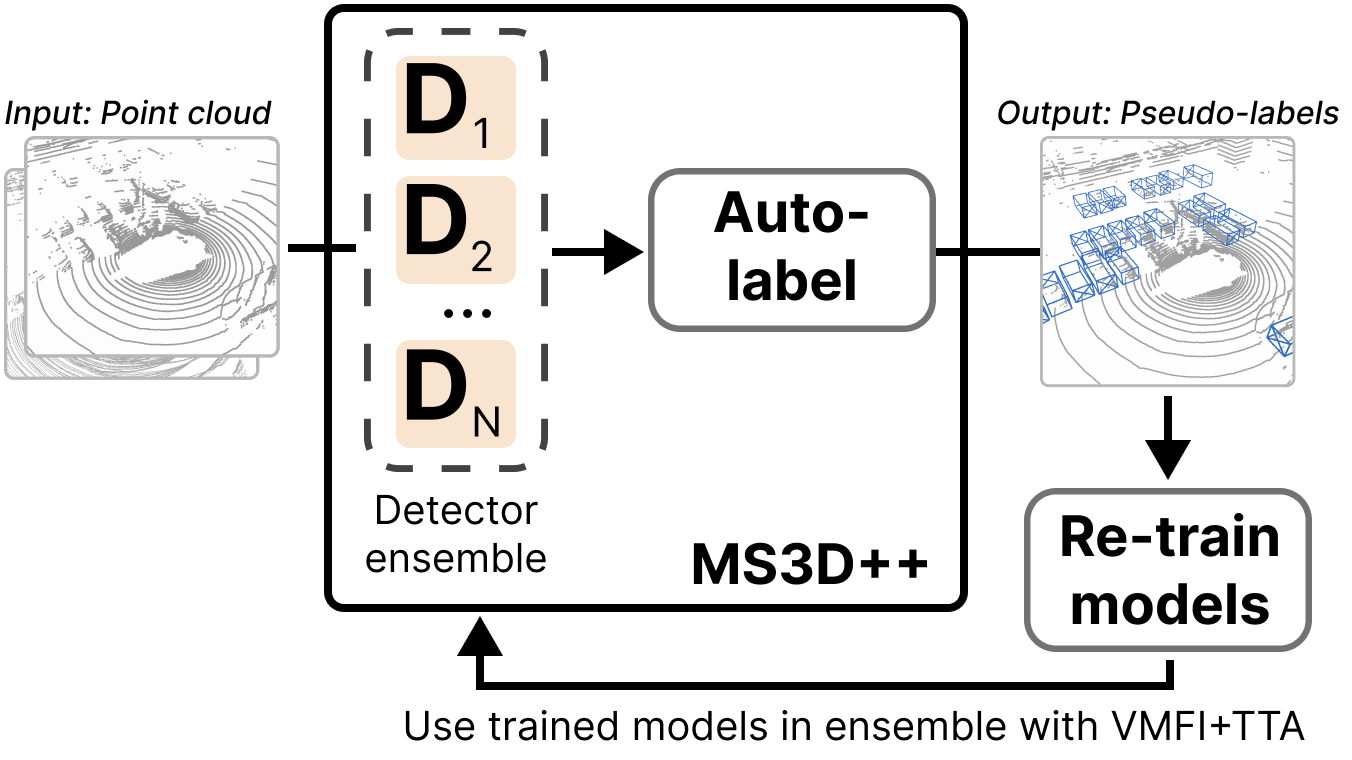}
  \caption{Multi-stage training process for MS3D++: In the first round of self-training, we employ a range of detectors from various source domains. The pseudo-labels are used to re-train the 3D detectors with multi-frame accumulation. We use the re-trained detectors to create a new ensemble for the next round.}
  \label{fig:multi_stage_framework}
  \vspace{-4mm}
\end{figure}

\subsection{Multi-Stage Self-Training}
\label{sec:multi_stage_self_training}
Iterative improvement through multiple self-training rounds has been shown to be a promising technique \cite{you2022modest,zhang2023oyster}. With each self-training round, the re-trained detector has increased detection range and improved alignment between score and class/localization. This increases the confidence scores for true predictions allowing easy distinction from false positives. The improvement in score alignment also extends to higher confidence detections at farther ranges, allowing for more precise tracking and refinement of far-range static vehicles and dynamic pedestrians for the subsequent pseudo-label set. The multi-stage self-training for MS3D++ is shown in \cref{fig:multi_stage_framework}. For our first round of self-training, we use a large ensemble of pre-trained detectors from various source domains for pseudo-label generation. With this pseudo-label set, we re-train a set of detectors with short-sequence, multi-frame accumulation. At this stage, VMFI is not required as the re-trained models are trained on the target domain's multi-frame scan pattern. To mitigate confirmation bias (i.e., the reinforcement of false positives) during each round of self-training, we set high $s_\text{pos}$ and $N_\text{pos}$ thresholds for all classes. Given the ego-vehicle motion in the scene, driving past tracked objects often result in a few high confidence detections at some point in the driving sequence. This in turn enables our static vehicle and dynamic pedestrian refinement to extrapolate and interpolate pseudo-labels to all tracked observations thereby increasing recall. This conservative approach for selecting $s_\text{pos}$ and $N_\text{pos}$ enables the gradual increase of pseudo-label quality and recall while maintaining high precision. 


\begin{figure*}
  \centering
  \includegraphics[width=0.99\linewidth]{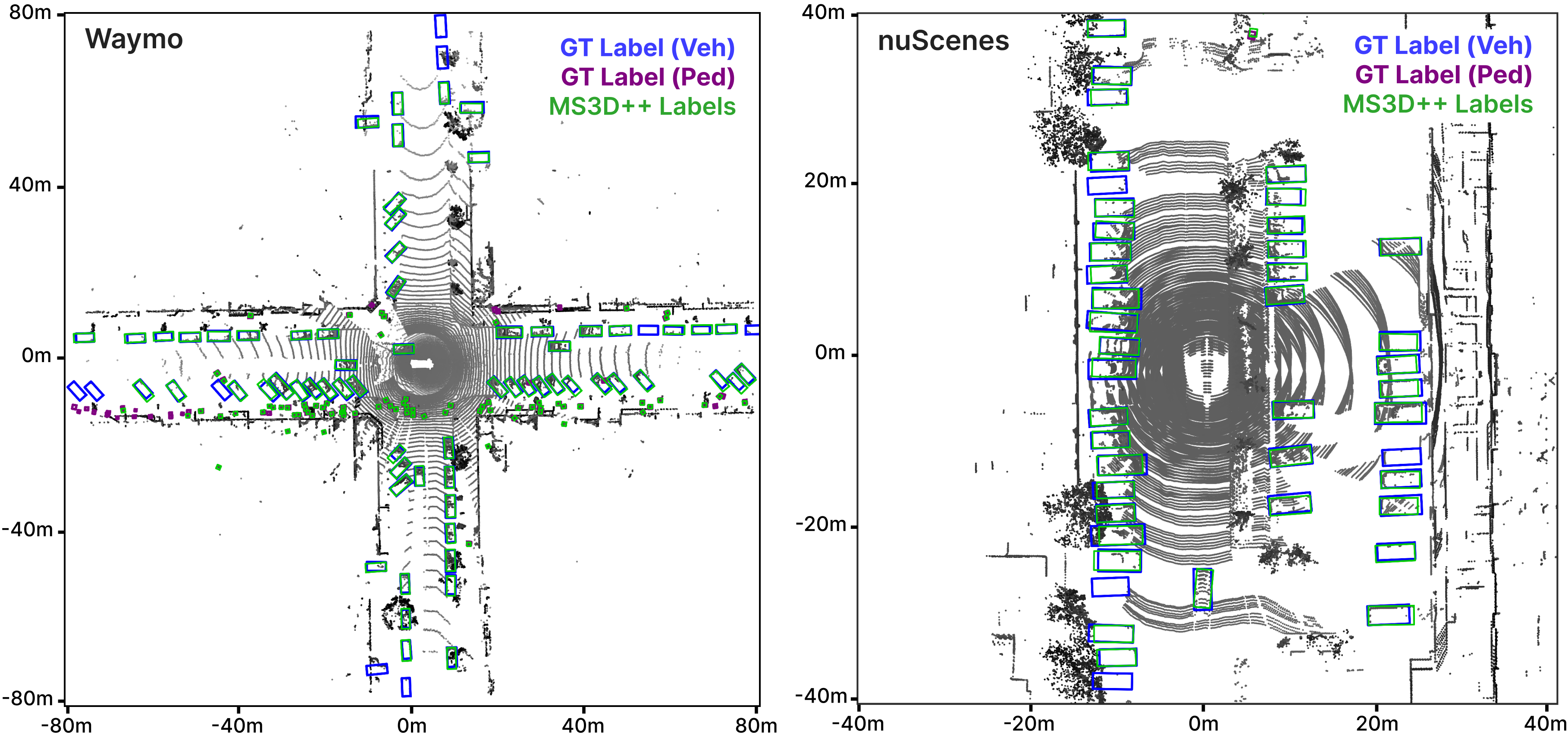}
  \caption{Visualization of MS3D++ pseudo-labels on Waymo and nuScenes point clouds. We show that we can obtain well localized boxes for sparse observations on both datasets. For a 64-beam lidar (Waymo), we show that MS3D++ can generate accurate pseudo-labels for up to 80m range.}
  \label{fig:ms3d_auto_labels}
  \vspace{-2mm}
\end{figure*}

\section{Experiments}
\label{sec:experiments}
\subsection{Setup}
\subsubsection{Datasets and evaluation metrics} For our experiments we use three datasets: Waymo, nuScenes and Lyft, as introduced in \cref{sec:quantifying_cross_domain_detection} where each large-scale dataset are varied in lidar scan pattern, weather and geography. The datasets are summarised in \cref{tab:datasets}. Due to label inconsistency in the cyclist category across domains as mentioned in \cref{sec:study_cross_domain_results}, we focus on the ``Vehicle" and ``Pedestrian" class. For nuScenes and Lyft, our ``Vehicle" supercategory includes the car, truck and bus subcategories. Following existing works \cite{yang2021st3d,hu2023density,wei2022lidardistillation}, we use KITTI average precision metric over 40 recall positions in BEV and 3D, denoted as $\text{AP}_\text{BEV}$ and $\text{AP}_\text{3D}$. For all tables we report AP on the official validation split of each dataset at IoU thresholds of 0.7 for ``Vehicle" and 0.5 for ``Pedestrian".

\subsubsection{Comparison methods} We compare MS3D++ with the following methods: (1) Source-only, directly applying the detector on the target domain without any domain adaptation; (2) ST3D \cite{yang2021st3d} serves as a single-source, self-training baseline, using a single detector without temporal refinement for pseudo-labeling; (3) MS3D \cite{tsai2023ms3d}, our previous work which used SECOND and Centerpoint for detector ensembling with single-stage self-training; (4) GT-trained, uses manually annotated ground truth labels to train the detector, providing an upper bound comparison if we were to achieve perfect pseudo-labels. For nuScenes as target domain, we further include other methods that report results for the Waymo$\rightarrow$nuScenes domain adaptation task: LD (with ST3D) \cite{wei2022lidardistillation}, DTS \cite{hu2023density} and REDB \cite{chen2023revisiting}. Similar to \cite{tsai2023ms3d}, GT-trained presents an apples-to-apples comparison of our pseudo-labels to GT labels. Between GT-trained and MS3Ds, the following are kept the same: data augmentation, dataset size, voxel size, anchor box sizes, number of epochs, batch size and learning rate. This comparison contrasts with the typically reported upper bound ``Oracle" supervised benchmark with optimized hyperparameters \cite{yang2021st3d,hu2023density} . For all methods, we train with 4-frame accumulation on Waymo ($T_{\Delta \text{MAX}}=0.3$), 3-frame on Lyft ($T_{\Delta \text{MAX}}=0.4$), and 10-frame on nuScenes ($T_{\Delta \text{MAX}}=0.45$) unless specified otherwise in the table.

\subsubsection{Implementation details}
We use the pre-trained detectors: PV-RCNN Anchorhead (PV-A), PV-RCNN Centerhead (PV-C), VoxelRCNN Anchorhead (VX-A) and VoxelRCNN Centerhead (VX-C) and their training configurations as specified in \cref{sec:quantifying_cross_domain_exp_setup}. Similar to \cite{tsai2023ms3d}, we sort Waymo and nuScenes by number of vehicles and pedestrians per scene, selecting the top 190 and 193 scenes, respectively. This selection allows us to focus on the static vehicle refinement, as scenes with many cars often represent parking lots. For an unlabeled target domain, a similar approach can be adopted by intentionally collecting data from scenes with carparks and densely populated pedestrian areas. The initial round of pseudo-labels are generated with four pre-trained detectors from various source domains and used to train VX-A and VX-C with multi-frame accumulation. For subsequent self-training rounds, we only use VX-A and VX-C in the ensemble with TTA. This choice is due to the distinct box proposal approaches of anchorhead \cite{yan2018second,he2017maskrcnn} and centerhead \cite{yin2021centerpoint}, which provide more varied object proposals. For pre-trained detector KBF weighing on Waymo as the target domain, we assign higher weights to Lyft detectors over nuScenes detectors due to Lyft having 64-beam point clouds. Conversely, when Lyft is the target domain, we assign higher weights to Waymo detectors compared to nuScenes detectors. Lastly, for nuScenes as the target domain, we weigh Waymo and Lyft detectors equally. Further implementation details can be found in our open-sourced code.

{\renewcommand{\arraystretch}{1.2}%
\begin{table}[]
\centering
\setlength{\tabcolsep}{0.3em} 
\caption{\textbf{Waymo} as the target domain for unsupervised domain adaptation. Detectors for all methods are trained on 4-frame accumulation ($T_{\Delta \text{MAX}}=0.3$). }
\label{tab:main_results_waymo}
\begin{tabular}{c|c|c|cc|cc}
\hline
\multirow{2}{*}{Method} & \multirow{2}{*}{Label Source} & \multirow{2}{*}{Detector} & \multicolumn{2}{c|}{Vehicle} & \multicolumn{2}{c}{Pedestrian} \\
                        &                               &                           & $\text{AP}_\text{BEV}$           & $\text{AP}_\text{3D}$           & $\text{AP}_\text{BEV}$            & $\text{AP}_\text{3D}$            \\ \hline
Source-Only             & Lyft                          & VX-C                     & 58.34         & 44.96        & 41.46          & 36.30         \\
Source-Only             & Lyft                          & VX-A                     & 59.06         & 45.99        & 39.56          & 34.04         \\ \hline\hline
ST3D \cite{yang2021st3d}                    & Lyft                          & VX-C                     & 57.43         & 43.68        & 45.02          & 41.62         \\
MS3D \cite{tsai2023ms3d}                    & Lyft,nuScenes                & VX-C                     & 64.34         & 47.75        & -              & -             \\ \hdashline[7pt/2pt]
MS3D++                  & Lyft,nuScenes                & VX-C                     & \textbf{70.60}         & \textbf{52.80}        & \textbf{57.04}          & \textbf{51.83}         \\
MS3D++                  & Lyft,nuScenes                & VX-A                     & \textbf{70.31}         & \textbf{52.31}        & \textbf{52.71}          & \textbf{48.92}         \\ \hline\hline
GT-Trained              & Waymo                             & VX-C                     & 75.11         & 61.28        & 67.86          & 62.92         \\
GT-Trained              & Waymo                             & VX-A                     & 73.81         & 60.52        & 57.89          & 54.71         \\ \hline
\end{tabular}
\vspace{-2mm}
\end{table}}

\subsection{Results}
In \cref{tab:main_results_waymo,tab:main_results_nusc}, we demonstrate that training VX-A and VX-C with MS3D++ pseudo-labels yields an $\text{AP}_\text{BEV}$ that outperforms all existing works and achieves results comparable to training with manually labeled ground-truths for Waymo and nuScenes datasets. \cref{fig:ms3d_auto_labels} shows examples of the labels generated with MS3D++ used in training. For Waymo as the target domain, the $\text{AP}_\text{BEV}$ difference between training VX-A with MS3D++ labels compared to GT labels is 3.5 $\text{AP}_\text{BEV}$ for ``Vehicle" and 5.17 $\text{AP}_\text{BEV}$ for ``Pedestrian". Similarly, for nuScenes, VX-A trained with MS3D++ shows only a 3.28 $\text{AP}_\text{BEV}$ reduction compared to training with GT labels. Additionally, MS3D++ outperforms all existing works in $\text{AP}_\text{3D}$ for Waymo but is around $-1 \text{AP}_\text{3D}$ lower in nuScenes. One reason for this is due to the lidar sparsity of nuScenes where poor box height localization for a sparse observation can lead to the pseudo-label containing more ground points than object points, therefore affecting label quality. 

{\renewcommand{\arraystretch}{1.2}%
\begin{table}[]
\centering
\setlength{\tabcolsep}{0.1em} 
\caption{\textbf{nuScenes} as the target domain for unsupervised domain adaptation. Sweeps refers to the number of accumulated frames that the detector was trained on.}
\label{tab:main_results_nusc}
\begin{tabular}{c|c|c|c|cc|cc}
\hline
\multirow{2}{*}{Method} & \multirow{2}{*}{Sweeps} & \multirow{2}{*}{Label Source} & \multirow{2}{*}{Detector} & \multicolumn{2}{c|}{Vehicle} & \multicolumn{2}{c}{Pedestrian} \\
                        &                         &                               &                           & $\text{AP}_\text{BEV}$            & $\text{AP}_\text{3D}$            & $\text{AP}_\text{BEV}$             & $\text{AP}_\text{3D}$            \\ \hline
Source-Only             & 4                       & Waymo                         & VX-C                      & 42.18         & 22.29        & 16.30          & 11.02         \\
Source-Only             & 4                       & Waymo                         & VX-A                      & 40.53         & 21.17        & 13.19          & 9.81          \\
Source-Only             & 1                       & Waymo                         & SECOND                    & 32.10         & 17.77        & 7.32              & 5.01             \\ \hline\hline
REDB \cite{chen2023revisiting}                    & 1                       & Waymo                         & SECOND                    & 30.12         & 18.56        & 2.47              & 2.14             \\
ST3D \cite{yang2021st3d}                    & 1                       & Waymo                         & SECOND                    & 35.90         & 20.20        & 5.75           & 5.11          \\

DTS \cite{hu2023density}                    & 1                       & Waymo                         & SECOND                    & 41.20         & 23.00        & -              & -             \\
L.D \cite{wei2022lidardistillation}                    & 1                       & Waymo                         & SECOND                    & 42.04         & 24.50        & -              & -             \\
MS3D \cite{tsai2023ms3d}                    & 1                       & Lyft, Waymo                   & SECOND                    & 42.23         & \textbf{24.76}        & -              & -             \\ \hdashline[7pt/2pt]
MS3D++                  & 1                       & Lyft, Waymo                   & SECOND                    & \textbf{43.88}         & 23.10        & -              & -             \\ \hline
ST3D \cite{yang2021st3d}                    & 10                      & Waymo                         & VX-C                      & 38.80         & 23.37        & 11.59          & 9.04          \\
MS3D \cite{tsai2023ms3d}                    & 10                      & Lyft, Waymo                   & VX-C                      & 49.23         & \textbf{27.50}        & -              & -             \\ \hdashline[7pt/2pt]
MS3D++                  & 10                      & Lyft, Waymo                   & VX-C                      & \textbf{50.34}         & 27.20        & \textbf{25.81}          & \textbf{15.91}         \\
MS3D++                  & 10                      & Lyft, Waymo                   & VX-A                      & \textbf{52.06}         & \textbf{26.52}        & \textbf{27.01}          & \textbf{15.36}         \\ \hline\hline
GT-Trained              & 1                       & nuScenes                      & SECOND                    & 44.39         & 29.46        & -              & -             \\
GT-Trained              & 10                      & nuScenes                      & VX-C                      & 56.43         & 37.26        & 41.77          & 32.52         \\
GT-Trained              & 10                      & nuScenes                      & VX-A                      & 55.34         & 36.67        & 38.30          & 29.72         \\ \hline
\end{tabular}
\vspace{-4mm}
\end{table}}

Overall, the $\text{AP}_\text{3D}$ lags behind $\text{AP}_\text{BEV}$ as the label quality is heavily impacted by the challenging nature of box height estimation for sparsely observed objects on all datasets as shown in \cref{fig:dataset_domain_gap}. Nonetheless, we highlight that BEV is often sufficient for various autonomous driving tasks such as obstacle avoidance, motion forecasting, and path planning. We believe that further improvement in 3D box localization can be achieved through leveling the pseudo-label boxes with ground plane segmentation \cite{oh2022travel,lee2022patchworkpp} of all accumulated sequence frames, or using a 3D map \cite{zhang2014loam}. Additionally, we notice that training a single multi-frame detector on its own pseudo-labels (ST3D) can lead to a degradation in performance compared to the Source-only approach for all three target domains. This can be attributed to an amplified domain gap in the multi-frame setting. The gap is likely caused by inadequate localization of dynamic objects, stemming from significant differences in the appearances of motion tails and accumulated point clouds across datasets. Hence, although ST3D may be applicable in a multi-frame adaptation setting, its lack of accurate pseudo-labels ultimately leads to unsatisfactory domain transfer for these datasets.

\begin{figure}[t]
  \centering
  \includegraphics[width=0.95\linewidth]{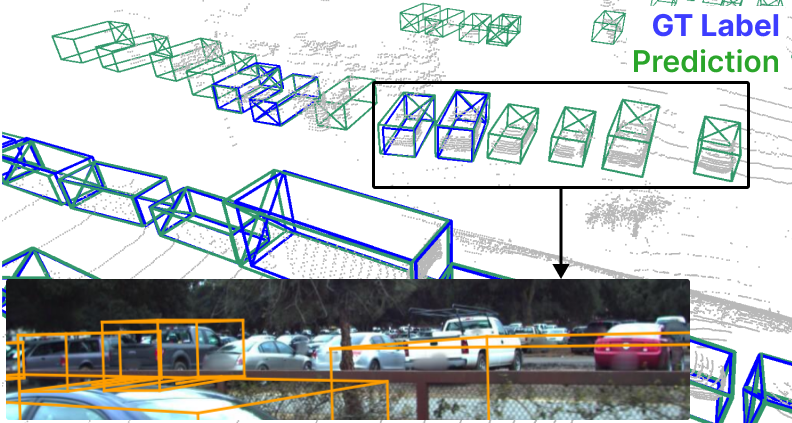}
  \caption{Missing labels on the Lyft dataset affects accurate evaluation of domain adaptation performance. Additionally, this contributes to the poor performance when testing Lyft-trained detectors on Waymo or nuScenes due to the difference in labeling guidelines.}
  \label{fig:lyft_missing_labels}
  \vspace{-3mm}
\end{figure}

For Lyft as the target domain in \cref{tab:main_results_lyft}, we obtain a higher $\text{AP}_\text{3D}$ than all existing works. However, similar to the observation made by \cite{yang2021st3d}, we illustrate in \cref{fig:lyft_missing_labels} that Lyft has many missing ground-truth labels for parked vehicles on the side of the road, as their primary focus is on motion prediction of traffic participants \cite{houston2020lyftmotion}. This makes it challenging to truly evaluate the domain adaptation performance of all methods due to predictions being penalized as false positives. Despite this limitation, we include these results from one round of self-training for the sake of completeness in our evaluation. We postulate that this labeling discrepancy plays a significant role in why Lyft-trained detectors often overlook many densely observed vehicles in Waymo and nuScenes, resulting in a notable decline in performance.

{\renewcommand{\arraystretch}{1.2}%
\begin{table}[]
\centering
\setlength{\tabcolsep}{0.15em} 
\caption{\textbf{Lyft} as the target domain for unsupervised domain adaptation. Detectors for all methods are trained on 3-frame accumulation ($T_{\Delta \text{MAX}}=0.4$). }
\label{tab:main_results_lyft}
\begin{tabular}{c|c|c|cc|cc}
\hline
\multirow{2}{*}{Method} & \multirow{2}{*}{Label Source} & \multirow{2}{*}{Detector} & \multicolumn{2}{c|}{Vehicle}    & \multicolumn{2}{c}{Pedestrian}  \\
                        &                               &                           & $\text{AP}_\text{BEV}$            & $\text{AP}_\text{3D}$             & $\text{AP}_\text{BEV}$            & $\text{AP}_\text{3D}$             \\ \hline
Source-Only             & Waymo                         & VX-C                     & 72.09          & 57.55          & 43.74          & 38.42          \\
Source-Only             & Waymo                         & VX-A                     & 74.28          & 59.69          & 42.52          & 37.87          \\ \hline\hline
ST3D \cite{yang2021st3d}                    & Waymo                         & VX-C                     & 70.57          & 57.08          & \textbf{47.25} & 41.84          \\
MS3D \cite{tsai2023ms3d}                    & Waymo,nuScenes               & VX-C                     & \textbf{77.30} & 63.43          & -              & -              \\ \hdashline[7pt/2pt]
MS3D++                  & Waymo,nuScenes               & VX-C                     & 77.00          & \textbf{65.96} & 46.89          & \textbf{43.26} \\
MS3D++                  & Waymo,nuScenes               & VX-A                     & \textbf{77.20} & \textbf{65.30} & \textbf{47.15} & \textbf{43.62} \\ \hline\hline
GT-Trained              & Lyft                             & VX-C                     & 86.85          & 74.76          & 60.67          & 54.18          \\
GT-Trained              & Lyft                             & VX-A                     & 85.23          & 72.70          & 58.86          & 50.37          \\ \hline
\end{tabular}
\end{table}}

A major advantage of MS3D++ lies in the flexibility of simply having a set of pseudo-labels for domain adaptation. This means that data augmentation techniques that proves effective in a supervised setting such as multi-frame accumulation can seamlessly integrate with our approach. From our results, we demonstrate the successful application of multi-frame accumulation for multiple classes across various target domains, yielding a significant improvement over Source-only for both rigid and non-rigid (i.e., ``Vehicle" and ``Pedestrian") classes on multiple target domains. Likewise, we believe that further improvements can be obtained through the use of the popular ground-truth database data augmentation \cite{yan2018second} to increase the number of labeled samples per point cloud for each round of self-training or just the final round.

\subsection{Ablation}
\label{sec:ablation}

\subsubsection{VMFI} 
We validate VMFI's effectiveness in \cref{tab:ablation_vmfi} using two models and source domains, showing its consistent improvement of precision and recall across all classes and ranges. For vehicles, VMFI significantly improves recall at all ranges, mainly due to increased point density for better detection of static and slow-moving vehicles. Meanwhile, for pedestrians, VMFI substantially increases precision across all ranges, achieving an improvement of up to 20\%. This improvement is due to the fact that pedestrians have sparser observations compared to vehicles, primarily due to their smaller size. Multi-frame accumulation is particularly effective on pedestrian detection due to their slow movement speed, resulting in better box localization from increased point density, with minimal side effects from long motion tails. Additionally, incorporating TTA further enhances both precision and recall. 

{\renewcommand{\arraystretch}{1.1}%
\begin{table}
\centering
\setlength{\tabcolsep}{0.15em} 
\caption{VMFI and TTA for ensembling. Precision (P) and Recall (R) range breakdown on Waymo point clouds. Source is given as Training dataset / $T_{\Delta \text{MAX}}$ / Detector.}
\label{tab:ablation_vmfi}
\begin{tabular}{c|c|cc|cc|cc|cc}
\hline
\multirow{2}{*}{Source}     & \multirow{2}{*}{Class}      & \multirow{2}{*}{VMFI} & \multirow{2}{*}{TTA} & \multicolumn{2}{c|}{0-30m}      & \multicolumn{2}{c|}{30-50m}     & \multicolumn{2}{c}{50-80m}      \\
                            &                             &                       &                      & P              & R              & P              & R              & P              & R              \\ \hline
\multirow{6}{*}{L / 0.4 / VX-C}  & \multirow{3}{*}{Veh.}    &                       &                      & 0.960          & 0.770          & 0.911          & 0.455          & 0.860          & 0.208          \\
                            &                             & \checkmark                  &                      & 0.959          & 0.805          & 0.919          & 0.492          & 0.868          & 0.233          \\
                            &                             & \checkmark                  & \checkmark                 & \textbf{0.962} & \textbf{0.820} & \textbf{0.926} & \textbf{0.508} & \textbf{0.886} & \textbf{0.242} \\ \cline{2-10} 
                            & \multirow{3}{*}{Ped.} &                       &                      & 0.726          & 0.425          & 0.746          & 0.298          & 0.771          & 0.143          \\
                            &                             & \checkmark                  &                      & 0.868          & 0.444          & 0.851          & 0.315          & 0.838          & 0.154          \\
                            &                             & \checkmark                  & \checkmark                 & \textbf{0.872} & \textbf{0.489} & \textbf{0.867} & \textbf{0.342} & \textbf{0.863} & \textbf{0.157} \\ \hline
\multirow{6}{*}{N / 0.45 / PV-C} & \multirow{3}{*}{Veh.}    &                       &                      & 0.976          & 0.693          & 0.957          & 0.318          & 0.949          & 0.051          \\
                            &                             & \checkmark                  &                      & \textbf{0.976} & 0.769          & 0.957          & 0.377          & 0.952          & 0.067          \\
                            &                             & \checkmark                  & \checkmark                 & 0.975          & \textbf{0.805} & \textbf{0.958} & \textbf{0.418} & \textbf{0.957} & \textbf{0.078} \\ \cline{2-10} 
                            & \multirow{3}{*}{Ped.} &                       &                      & 0.533          & 0.194          & 0.535          & \textbf{0.038} & 0.625          & \textless 0.01 \\
                            &                             & \checkmark                  &                      & 0.652          & 0.184          & 0.653          & 0.027          & \textbf{0.690} & \textless 0.01 \\
                            &                             & \checkmark                  & \checkmark                 & \textbf{0.671} & \textbf{0.205} & \textbf{0.680} & 0.029          & 0.500          & \textless 0.01 \\ \hline
\end{tabular}
\end{table}}

\begin{table}[]
\centering
\setlength{\tabcolsep}{0.1em} 
\caption{Ensemble combinations for Lyft (L) and nuScenes (N) detectors on Waymo point clouds. We report $\text{AP}_\text{3D}$ in the ranges 0-30m, 30-50m and 50-80m.}
\label{tab:ensemble_combinations_kbf}
\begin{tabular}{c|cccc|ccc|ccc}
\hline
\multirow{2}{*}{Source} & \multirow{2}{*}{PV-A} & \multirow{2}{*}{PV-C} & \multirow{2}{*}{VX-A} & \multirow{2}{*}{VX-C} & \multicolumn{3}{c|}{Vehicle}                  & \multicolumn{3}{c}{Pedestrian}                \\
                        &                       &                       &                        &                        & 0-30          & 30-50         & 50-80         & 0-30          & 30-50         & 50-80         \\ \hline
L                       & \checkmark                  &                       &                        &                        & 74.4          & 44.4          & 20.1          & 38.5          & 25.6          & 12.8          \\
L                       &                       & \checkmark                  &                        &                        & 73.0          & 43.2          & 19.1          & 33.0          & 25.1          & 15.0          \\
L                       &                       &                       & \checkmark                   &                        & 74.6          & 46.0          & 21.4          & 43.6          & 32.0          & 20.5          \\
L                       &                       &                       &                        & \checkmark                   & 73.1          & 45.8          & 21.6          & 46.6          & 37.2          & 27.0          \\ \hline
L                       &                       &                       & \checkmark                   & \checkmark                   & 79.5          & 52.5          & 26.2          & \textbf{57.3} & 46.6          & \textbf{31.8} \\
L                       & \checkmark                  & \checkmark                  &                        &                        & 78.1          & 49.7          & 24.1          & 48.7          & 35.9          & 19.8          \\
L                       &                       & \checkmark                  &                        & \checkmark                   & 78.2          & 50.8          & 25.1          & 54.6          & 45.1          & 30.8          \\
L                       & \checkmark                  &                       & \checkmark                   &                        & 79.8          & 52.4          & 26.2          & 51.7          & 38.1          & 22.0          \\
L                       &                       & \checkmark                  & \checkmark                   &                        & 79.3          & 51.8          & 25.4          & 52.6          & 40.3          & 24.1          \\
L                       & \checkmark                  &                       &                        & \checkmark                   & 79.2          & 52.2          & 26.1          & 55.2          & 44.5          & 30.1          \\
L                       & \checkmark                  & \checkmark                  & \checkmark                   & \checkmark                   & 80.6          & 54.6          & \textbf{28.0} & 57.1          & 46.7          & 31.7          \\ \hline
N+L                     & \checkmark                  & \checkmark                  & \checkmark                   & \checkmark                   & 82.6          & 52.7          & 25.1          & 46.8          & 38.6          & 28.5          \\
$\text{N+L}_\text{weight}$         & \checkmark                  & \checkmark                  & \checkmark                   & \checkmark                   & \textbf{82.6} & \textbf{55.7} & 27.8          & 57.1          & \textbf{46.7} & 31.7          \\ \hline
\end{tabular}
\vspace{-2mm}
\end{table}

\subsubsection{Ensemble contributions and weighing} \label{sec:ablation_ensemble}
In \cref{tab:ensemble_combinations_kbf}, we present the performance of various detector combinations on the Waymo dataset. We demonstrate that any ensembling any pair of detectors with TTA and VMFI easily surpasses the performance of using a single detector for pseudo-labeling. Although the Lyft-trained VX-A and VX-C are the highest performing pre-trained detectors on Waymo, we show that combining all detectors with KBF outperforms the VX-A and VX-C pair. However, when introducing nuScenes detectors into the ensemble, we observed a decrease in overall pedestrian and 30-80m vehicle detection. nuScenes-trained detectors tends to predict more false positive pedestrians and vehicles, particularly at farther ranges. This is potentially due to the fact that nuScenes detectors are trained to make highly confident predictions from sparse observations (e.g., a single lidar ring) for objects beyond 30m range. In a cross-domain context, this leads to noisy predictions due to a different distribution of points between source and target domains, particularly for semi-occluded or far-range structures. Based on these insights, we assign a higher weight to Lyft-trained detectors for vehicle predictions, and we discard the pedestrian predictions from nuScenes-trained models to reduce false positives in our ensemble. We demonstrate that a simple weighing of Lyft detectors for KBF can enhance the overall vehicle $\text{AP}_\text{3D}$. Furthermore, we emphasize that this aspect of MS3D++ is not limited to the selected detectors in this paper. Depending on the use-case, incorporating detectors with varying voxel sizes, multi-modal capabilities, or class-specific detectors into the ensemble can further enhance the robustness of the initial pseudo-label set.
\subsubsection{Label breakdown}
To better understand the effect of KBF ensembling and label refinement, we analyse the precision and recall of our pseudo-labels in \cref{tab:label_breakdown}. For the vehicle class, KBF ensembling (Ensemble), can boost both precision and recall for all ranges. However, when it comes to pedestrians, their small size makes them more prone to occlusion, resulting in sparser observations and poorer box localization. As a result, the precision for pedestrians is affected. Nonetheless, we find that ensembling generally improves the recall of pedestrians. Subsequently, we demonstrate that the final set of pseudo-labels after temporal refinement (Refined) greatly increases the recall of our ensemble with a slight trade-off in precision. The improvement in recall is most prominent at the 30-80m range, which effectively extends the detection range of detectors adapted to new target domains.

{\renewcommand{\arraystretch}{1.1}%
\begin{table}[]
\centering
\setlength{\tabcolsep}{0.4em} 
\caption{Label breakdown of KBF and temporal refinement in MS3D++. Precision (P) and Recall (R) breakdown by range for Waymo point clouds. Delta $\Delta$ is the difference between Ensemble and Refined.}
\label{tab:label_breakdown}
\begin{tabular}{c|c|cc|cc|cc}
\hline
                             &                         & \multicolumn{2}{c|}{0-30m}                     & \multicolumn{2}{c|}{30-50m}                    & \multicolumn{2}{c}{50-80m}                     \\
\multirow{-2}{*}{Class}      & \multirow{-2}{*}{Label} & P      & R                                     & P      & R                                     & P      & R                                     \\ \hline
                             & VX-A                    & 0.946  & 0.899                                 & 0.894  & 0.705                                 & 0.847  & 0.427                                 \\
                             & VX-C                    & 0.950  & 0.896                                 & 0.904  & 0.695                                 & 0.864  & 0.403                                 \\
                             & Ensemble                & 0.950  & 0.899                                 & 0.904  & 0.704                                 & 0.866  & 0.418                                 \\
                             & Refined                 & 0.944  & 0.912                                 & 0.893  & 0.763                                 & 0.851  & 0.514                                 \\ \cline{2-8} 
\multirow{-5}{*}{Vehicle}    & $\Delta$   & -0.006 & {\color[HTML]{036400} \textbf{0.013}} & -0.011 & {\color[HTML]{036400} \textbf{0.059}} & -0.015 & {\color[HTML]{036400} \textbf{0.096}} \\ \hline
                             & VX-A                    & 0.875  & 0.623                                 & 0.846  & 0.493                                 & 0.834  & 0.325                                 \\
                             & VX-C                    & 0.881  & 0.609                                 & 0.863  & 0.485                                 & 0.867  & 0.326                                 \\
                             & Ensemble                & 0.872  & 0.646                                 & 0.853  & 0.518                                 & 0.849  & 0.351                                 \\
                             & Refined                 & 0.846  & 0.689                                 & 0.818  & 0.574                                 & 0.802  & 0.431                                 \\ \cline{2-8} 
\multirow{-5}{*}{Pedestrian} & $\Delta$   & -0.026 & {\color[HTML]{036400} \textbf{0.043}} & -0.035 & {\color[HTML]{036400} \textbf{0.056}} & -0.047 & {\color[HTML]{036400} \textbf{0.080}} \\ \hline
\end{tabular}
\begin{tablenotes}\footnotesize
    \item [a] Ensemble (KBF): VX-C and VX-A with VMFI and TTA
    \item [b] Refined: final pseudo-label set of MS3D++
\end{tablenotes} 
\vspace{-4mm}
\end{table}}

\subsubsection{Multi-stage self-training} 
\begin{figure}[t]
  \centering
  \includegraphics[width=0.95\linewidth]{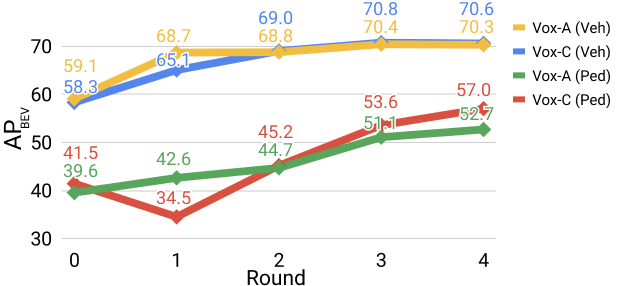}
  \caption{Multi-stage self-training on Waymo as the target domain with VoxelRCNN-Anchorhead (VX-A) and VoxelRCNN-Centerhead (VX-C). We show the improvement in detection rate of VX-A and VX-C over multiple rounds of self-training. All detectors are trained on 4-frame point cloud accumulation. The 0th round refers to the detectors without any domain adaptation (i.e., Source-Only).}
  \label{fig:multistage_ablation}
  \vspace{-4mm}
\end{figure}

\cref{fig:multistage_ablation} highlights the effectiveness of multiple rounds of self-training. The largest improvement in vehicle detection is in the 1st round of self-training using an ensemble of Lyft and nuScenes pre-trained detectors. However, as we proceed with more rounds of self-training, the improvement starts to plateau. On the other hand, for the pedestrian class, the 1st round of self-training leads to a decrease in $\text{AP}_\text{BEV}$ due to prioritizing a high precision for accurately distinguishing true pedestrians. Nevertheless, in subsequent rounds, more aligned confidence scores enables better thresholding of true pedestrians, allowing us to use more pedestrian tracks and increasing overall pseudo-label recall. We study this by examining the pseudo-label quality for each round in \cref{tab:multistage_ps_quality_ablation}. The analysis clearly demonstrates the high precision and low recall of the first round of pedestrian pseudo-labeling, with a substantial increase with each subsequent round. In general, for both classes, each round exhibits a significant increase in recall, with a minor trade-off in precision. However, we argue that for the autonomous driving application, a high recall is more important for safety considerations. The increase in recall is particularly pronounced at farther ranges (30-80m), where it is approximately 3 to 4 times larger than the drop in precision. For instance, the recall of the pedestrian class in round 4 at 50-80m is 10$\times$ larger than the recall in round 1, with only a 12\% decrease in precision. This is subsequently reflected in the significant improvement in pedestrian $\text{AP}_\text{BEV}$ in later self-training rounds. Similar to \cref{sec:ablation_ensemble}, the pseudo-labels for each round can be used to train any type of detector for the ensemble of subsequent rounds. This includes training class-specific or multi-modal detectors.

{\renewcommand{\arraystretch}{1.1}%
\begin{table}[]
\centering
\setlength{\tabcolsep}{0.5em} 
\caption{Multi-stage self-training. Precision (P) and Recall (R) range breakdown of the pseudo-label quality for each self-training round. Delta $\Delta$ is the difference between round 4 and round 1.}
\label{tab:multistage_ps_quality_ablation}
\begin{tabular}{c|c|cc|cc|cc}
\hline
                             &                         & \multicolumn{2}{c|}{0-30m}                     & \multicolumn{2}{c|}{30-50m}                    & \multicolumn{2}{c}{50-80m}                     \\
\multirow{-2}{*}{Class}      & \multirow{-2}{*}{Round} & P      & R                                     & P      & R                                     & P      & R                                     \\ \hline
                             & 1                       & 0.962  & 0.896                                 & 0.934  & 0.648                                 & 0.913  & 0.333                                 \\
                             & 2                       & 0.955  & 0.909                                 & 0.920  & 0.720                                 & 0.900  & 0.436                                 \\
                             & 3                       & 0.944  & 0.912                                 & 0.898  & 0.759                                 & 0.865  & 0.505                                 \\
                             & 4                       & 0.944  & 0.912                                 & 0.893  & 0.763                                 & 0.851  & 0.514                                 \\ \cline{2-8} 
\multirow{-5}{*}{Vehicle}    & $\Delta$              & -0.018 & {\color[HTML]{036400} \textbf{0.016}} & -0.041 & {\color[HTML]{036400} \textbf{0.115}} & -0.062 & {\color[HTML]{036400} \textbf{0.181}} \\ \hline
                             & 1                       & 0.912  & 0.324                                 & 0.922  & 0.131                                 & 0.906  & 0.043                                 \\
                             & 2                       & 0.881  & 0.532                                 & 0.891  & 0.372                                 & 0.909  & 0.182                                 \\
                             & 3                       & 0.874  & 0.628                                 & 0.861  & 0.497                                 & 0.870  & 0.322                                 \\
                             & 4                       & 0.846  & 0.689                                 & 0.818  & 0.574                                 & 0.802  & 0.431                                 \\ \cline{2-8} 
\multirow{-5}{*}{Pedestrian} & $\Delta$              & -0.066 & {\color[HTML]{036400} \textbf{0.365}} & -0.104 & {\color[HTML]{036400} \textbf{0.443}} & -0.104 & {\color[HTML]{036400} \textbf{0.388}} \\ \hline
\end{tabular}
\end{table}}

\subsubsection{Pseudo-label generation duration}
The first step of MS3D++ is to generate the predictions from each pre-trained detector with VMFI and TTA for the ensemble. The duration of this depends on the density of the target domain point clouds, detector architecture, and number of detectors used and can generally range from 3-12 hours. Following this, the remaining steps of MS3D++ can range from 6-20 hours depending on the number of objects in each frame. For instance, on 20k frames from the Waymo dataset, MS3D++ took around 20 hours due to our focus on carparks, and the extensive range of the 64-beam lidar. In contrast, a similar number of frames for nuScenes point clouds took around 6 hours as less objects are observed because of the limited usable range of a 32-beam lidar. A large bottleneck is from the tracker implementation \cite{pang2021simpletrack}, which accounts for up to 11 out of the 20 hours for the Waymo dataset. Overall, MS3D++ is still significantly more time efficient than manual labeling. We highlight that MS3D++ can also be used in conjunction with human annotators as a strong set of initial labels. MS3D++ only requires a GPU for pre-trained detector box proposals and detector training. The pseudo-labeling steps do not require a GPU for computation. Nevertheless, our experiments were run with GPU-optimized IoU computations on an i7-6700K CPU with 60GB RAM and a 2080TI GPU.

\section{Conclusion}
\label{sec:conclusion}
In this paper, we introduce MS3D++, a multi-source self-training framework that effectively generates high quality pseudo-labels for both low and high density lidar through detector ensembling and temporal refinement. Extensive experiments show that models trained with our pseudo-labels obtain state-of-the-art results on all evaluated target domains and are even comparable to training with manually annotated labels. Furthermore, our pseudo-labels are simply a drop-in replacement for ground-truth labels in a supervised training setting. This makes our domain adaptation approach highly versatile, supporting the training of newly developed 3D detectors, or data augmentation techniques. In summary, MS3D++ offers a practical and flexible solution for domain adaptation that maintains detector runtime and eliminates the need for modification of 3D detector architectures or custom learnable modules. Our future work is to extend our framework to incorporate multi-modal pre-trained detectors and active learning. This could involve quantifying the uncertainty of pseudo-labeled frames and replacing the pseudo-label of uncertain frames with human-annotated labels.








%
\bibliography{references}
\bibliographystyle{IEEEtran}

\begin{IEEEbiography}[{\includegraphics[width=1in,height=1.25in, clip,keepaspectratio]{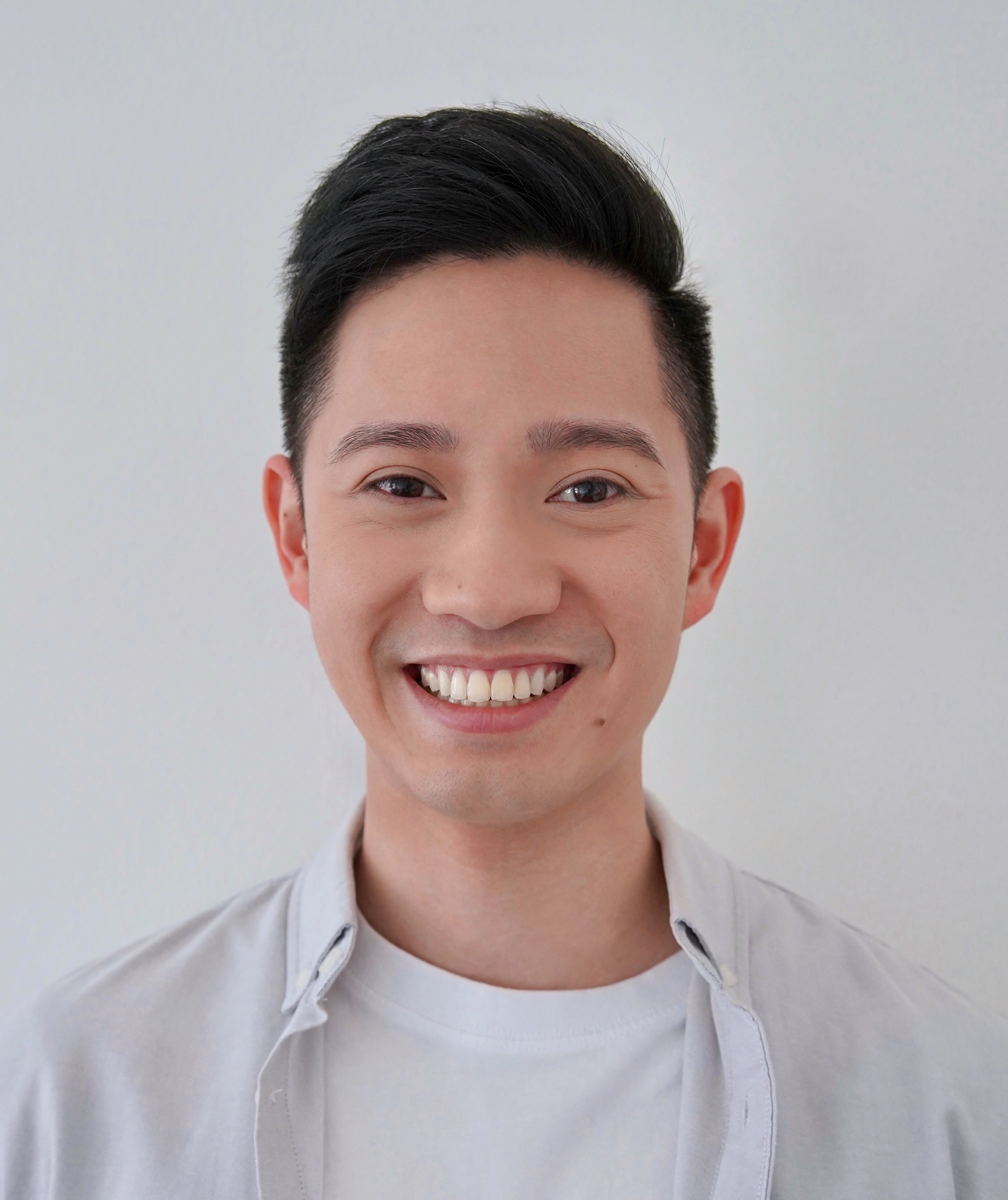}}]{Darren Tsai} received the B.S. degree in Electrical Engineering with First Class Honours from the University of Sydney, Australia, in 2020. He is currently working towards his Ph.D. degree at the University of Sydney. His research interests include computer vision and deep learning with a focus on unsupervised domain adaptation.
\end{IEEEbiography}
\vskip -2\baselineskip plus -1fil

\begin{IEEEbiography}[{\includegraphics[width=1in,height=1.25in, clip,keepaspectratio]{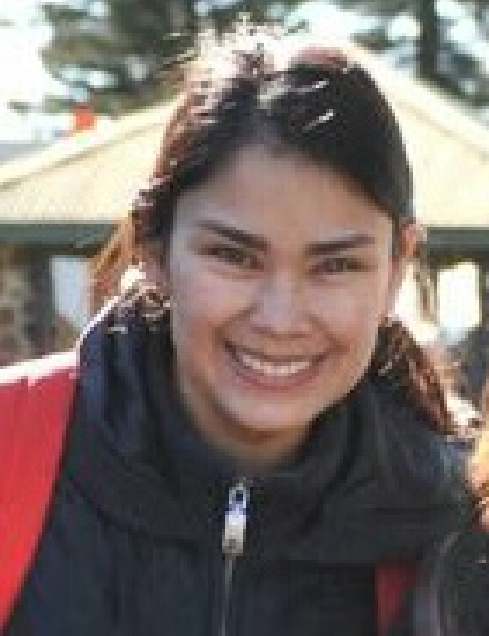}}]{Julie Stephany Berrio} received the B.S. degree in Mechatronics Engineering in 2009 from Universidad Autonoma de Occidente, Cali, Colombia, and the M.E. degree in 2012 from the Universidad del Valle, Cali, Colombia. She received her PhD from the University of Sydney, Australia, in 2021. She is a Research Associate with the Australian Centre for Robotics at the University of Sydney. Her research interest includes semantic mapping, machine learning and point cloud processing.
\end{IEEEbiography}
\vskip -2\baselineskip plus -1fil

\begin{IEEEbiography}[{\includegraphics[width=1in,height=1.25in,clip,keepaspectratio]{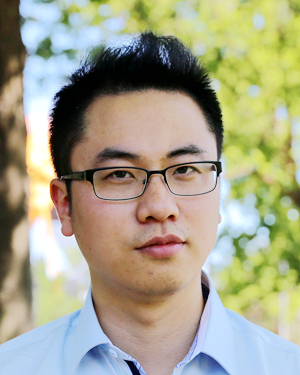}}]{Mao Shan} received the B.S. degree in electrical engineering from the Shaanxi University of Science and Technology, Xi’an, China, in 2006, and the M.S. degree in automation and manufacturing systems and PhD degree from the University of Sydney, Australia, in 2009 and 2014, respectively. He is a Research Fellow with the Australian Centre for Robotics at the University of Sydney. His research interests include autonomous systems, V2X, and tracking algorithms and applications.
\end{IEEEbiography}
\vskip -2\baselineskip plus -1fil

\begin{IEEEbiography}[{\includegraphics[width=1in,height=1.25in,clip,keepaspectratio]{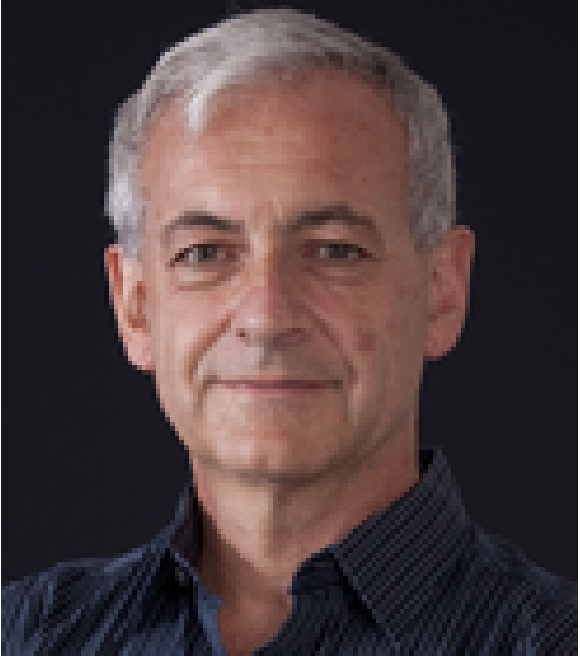}}]{Eduardo Nebot} received the BSc. degree in electrical engineering from the Universidad Nacional del Sur, Argentina, M.Sc. and PhD degrees from Colorado State University, Colorado, USA. He is currently an Emeritus Professor at the University of Sydney, Australia. His main research interests are in field robotics automation and intelligent transport systems. The major impact of his fundamental research is in autonomous systems, navigation, and safety.
\end{IEEEbiography}
\vskip -2.5\baselineskip plus -1fil

\begin{IEEEbiography}[{\includegraphics[width=1in,height=1.25in,clip,keepaspectratio]{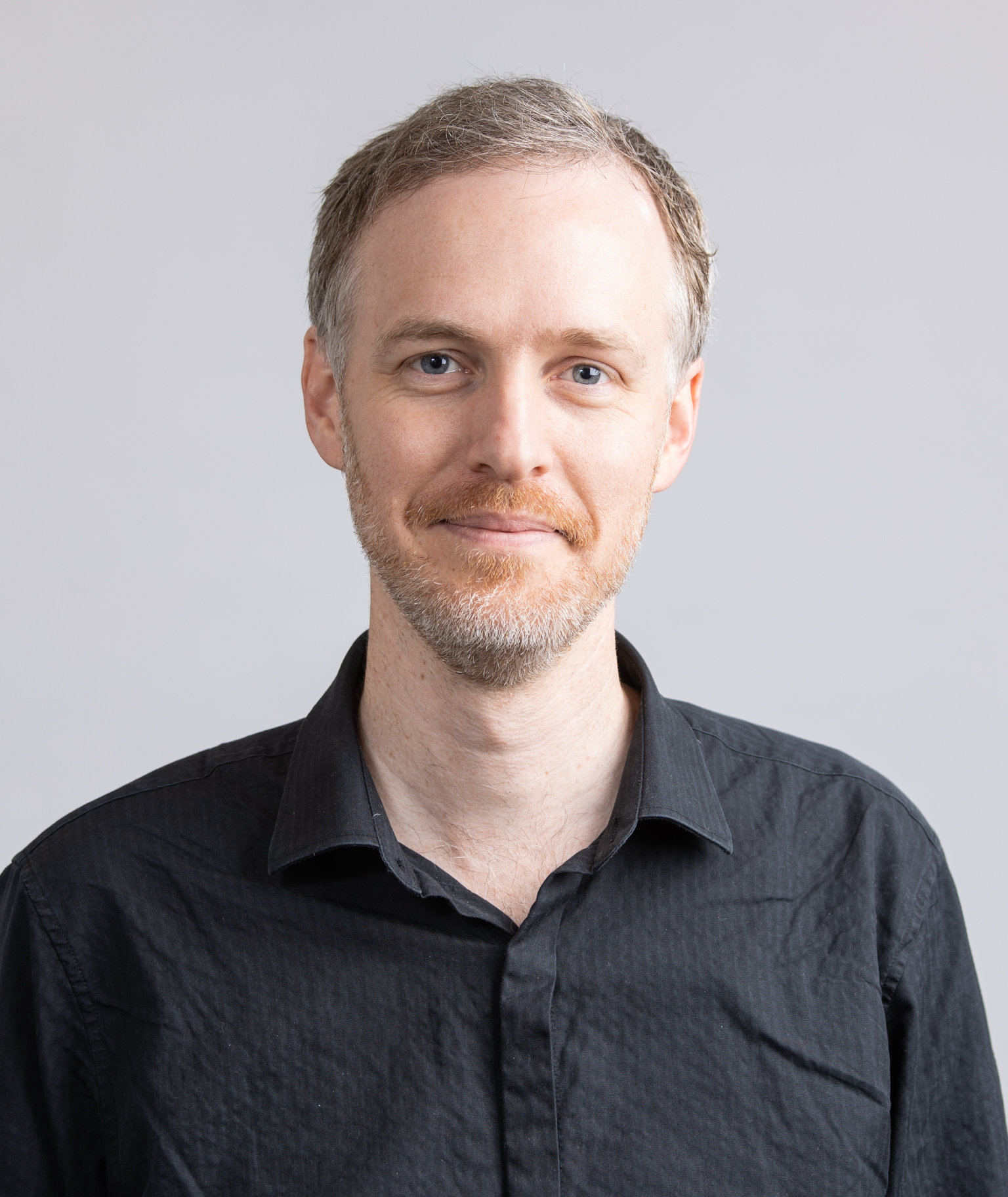}}]{Stewart Worrall} received his PhD from the University of Sydney, Australia, in 2009. He leads the Intelligent Transportation Systems group within the Australian Centre for Robotics at the University of Sydney. His research centers on improving technology for connected and autonomous vehicles to enhance safety and transform how people interact with these vehicles.
\end{IEEEbiography}

%








\end{document}